\RequirePackage{fix-cm}
\documentclass[twocolumn]{svjour3}          
\smartqed  
\usepackage{graphicx}%
\usepackage{multirow}%
\usepackage{amsmath,amssymb,amsfonts}%
\usepackage{mathrsfs}%
\usepackage[title]{appendix}%
\usepackage{xcolor}%
\usepackage{textcomp}%
\usepackage{manyfoot}%
\usepackage{booktabs}%
\usepackage{algorithm}%
\usepackage{algorithmicx}%
\usepackage{algpseudocode}%
\usepackage{listings}%
\usepackage[skins]{tcolorbox}
\usepackage[numbers]{natbib}
\usepackage{fbox}
\usepackage{makecell}
\usepackage{multirow}
\usepackage{inconsolata}
\usepackage{colortbl}
\usepackage{threeparttable}
\usepackage{footnote}
\usepackage{tablefootnote}
\usepackage{subfig,float}
\usepackage{color}
\usepackage[misc]{ifsym}
\usepackage{soul}
\usepackage{titlesec}
\usepackage{pifont}
\usepackage{utfsym}

\soulregister\cite7
\soulregister\ref7

\definecolor{darkergreen}{RGB}{21,152,56}
\definecolor{blue1}{RGB}{11,93,139}
\definecolor{yellow1}{RGB}{90,90,90}
\definecolor{skyblue}{RGB}{0,82,202}
\definecolor{dodgerblue}{RGB}{30,144,255}
\definecolor{lightblue}{RGB}{229,248,255}
\definecolor{green1}{rgb}{0.25,0.5,0.5}
\definecolor{red1}{rgb}{0.7,0.25,0.25}
\definecolor{gray1}{rgb}{0.85,0.85,0.85}
\definecolor{cvprblue}{rgb}{0.21,0.49,0.74}
\definecolor{colred}{RGB}{240,240,240}

\usepackage{paralist}

\makeatletter
\def\whline#1{%
	\noalign{\ifnum0=`}\fi\hrule \@height #1 \futurelet
	\reserved@a\@xhline}

\usepackage[colorlinks=true,linkcolor=skyblue,filecolor=skyblue,citecolor=skyblue]{hyperref}

\begin{document}

\title{Smaller But Better: Unifying Layout Generation with Smaller Large Language Models}

\titlerunning{Smaller But Better: Unifying Layout Generation with Smaller Large Language Models}


\author{
	Peirong Zhang\textsuperscript{1} \and 
    Jiaxin Zhang\textsuperscript{1} \and
    Jiahuan Cao\textsuperscript{1} \and
    Hongliang Li\textsuperscript{1} \and 
    Lianwen Jin\textsuperscript{1}
}


\institute{{\Letter}~Lianwen Jin (corresponding author)\vspace{-4mm} \\ 
\begin{quote}
	{eelwjin@scut.edu.cn}\vspace{-1.5mm}\\
	\\
	Peirong Zhang\\
	{eeprzhang@mail.scut.edu.cn}\vspace{-1.5mm}\\
	\\
	Jiaxin Zhang\\
	{msjxzhang@mail.scut.edu.cn}\vspace{-1.5mm}\\
	\\
	Jiahuan Cao\\
	{eejiahuancao@mail.scut.edu.cn}\vspace{-1.5mm}\\
	\\
	Hongliang Li\\
	{eehongliangli@mail.scut.edu.cn}\vspace{-1.5mm}\\
	\\
	\textsuperscript{1}\hspace{2pt}School of Electronic and Information Engineering, South China University of Technology, Guangzhou, China\vspace{-1.5mm}\\
	\end{quote}
}

\date{Received: 31 March 2024 / Accepted: 6 January 2025}

\maketitle

\begin{abstract}
We propose LGGPT, an LLM-based model tailored for unified layout generation. First, we propose Arbitrary Layout Instruction (ALI) and Universal Layout Response (ULR) as the uniform I/O template. ALI accommodates arbitrary layout generation task inputs across multiple layout domains, enabling LGGPT to unify both task-generic and domain-generic layout generation hitherto unexplored. Collectively, ALI and ULR boast a succinct structure that forgoes superfluous tokens typically found in existing HTML-based formats, facilitating efficient instruction tuning and boosting unified generation performance. In addition, we propose an Interval Quantization Encoding (IQE) strategy that compresses ALI into a more condensed structure. IQE precisely preserves valid layout clues while eliminating the less informative placeholders, facilitating LGGPT to capture complex and variable layout generation conditions during the unified training process. Experimental results demonstrate that LGGPT achieves superior or on par performance compared to existing methods. Notably, LGGPT strikes a prominent balance between proficiency and efficiency with a compact 1.5B parameter LLM, which beats prior 7B or 175B models even in the most extensive and challenging unified scenario. Furthermore, we underscore the necessity of employing LLMs for unified layout generation and suggest that 1.5B could be an optimal parameter size by comparing LLMs of varying scales. Code is available at \url{https://github.com/NiceRingNode/LGGPT}.

\end{abstract}

\section{Introduction}
\label{sec::intro}

\begin{figure}[b]
	\centering
	\includegraphics[width=\linewidth]{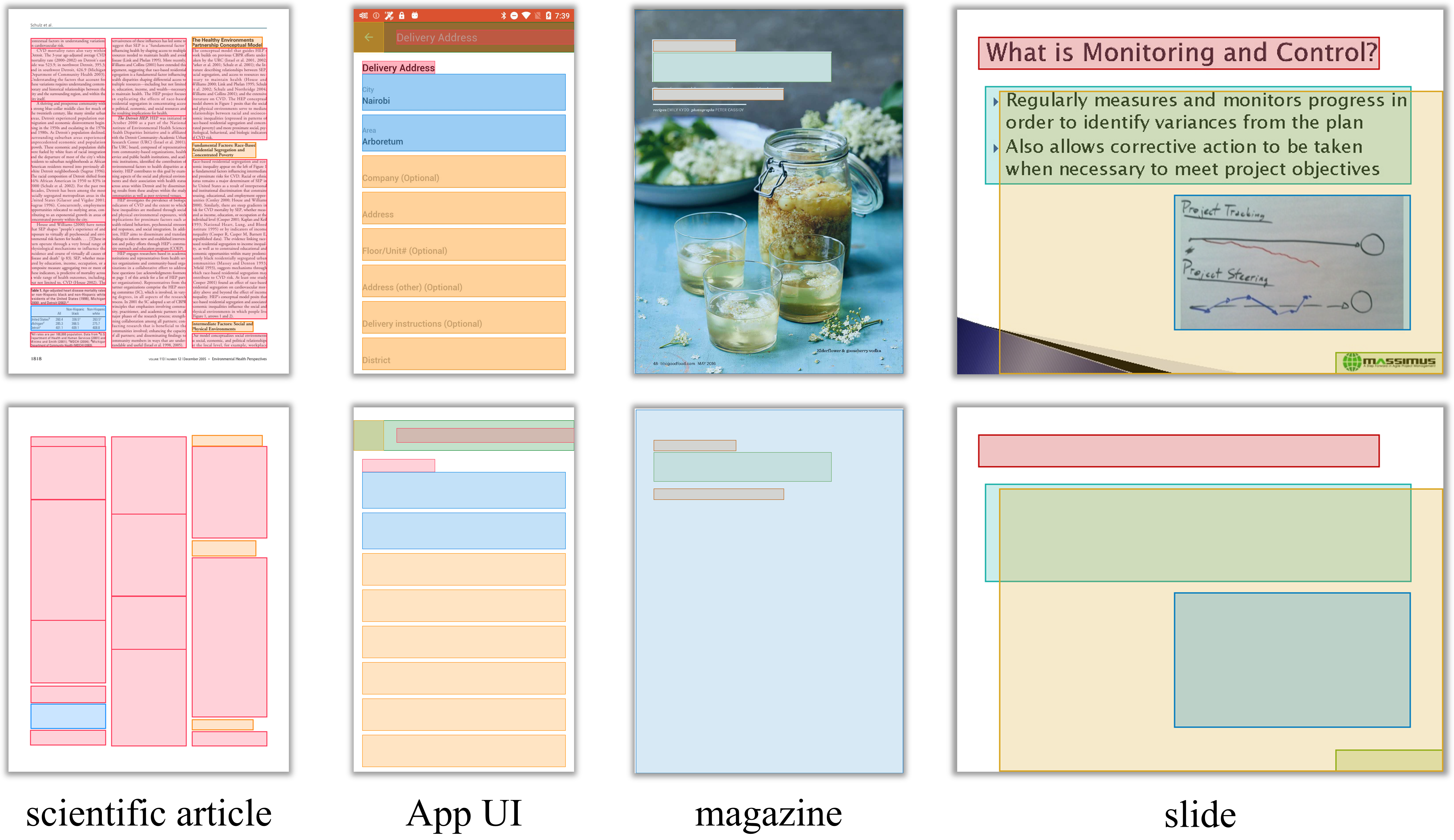}{}
	\caption{Examples of different types of layout.}
	\label{Fig::example}
\end{figure}

Graphic layout entails the structured arrangement of visual elements within a given space, as exemplified in Fig.~\ref{Fig::example}, playing a critical role in effective information display and visual perception. To circumvent manual design burdens, layout generation \cite{layoutganpp2021kikuchi,layoutdm2023Inoue,LDGM2023hui}, \emph{i.e.}, automatically creating realistic layouts tailored to diverse requirements, has fueled increasing fervor in the research community. Typically, layout is interpreted as a sequence of discrete element attributes, including element classes and geometric bounding boxes. Therefore, layout generation has been naturally framed as a sequence-to-sequence \cite{seq2seq2014nips} generation task.

Due to the inherent adaptability of Transformer \cite{attention2017vaswani} to sequential modeling, Transformer-based approaches \cite{layoutganpp2021kikuchi,vtn2021arroyo,ruite2021rahman,blt2022kong,layoutdm2023Inoue,LDGM2023hui,layouttfpp2023jiang,layoutdm2023chai,layoutdiff2023zhang,dlt2023levi,layoutprompter2023lin,layoutnuwa2024tang} have emerged as frontrunners in this field. The currently predominant paradigm is masked sequence modeling, which obfuscates certain element attributes to produce task-specific layout specifications as constraints. The model is required to predict the masked attributes in a fill-in-the-blank manner to accomplish layout generation. The other foundational paradigm is generative modeling using decoder-based Transformers, which generates new layouts in an auto-regressive manner. Unlike masked modeling, whose predicted values may lack internal logical coherence and result in chaotic layouts, especially with a high proportion of masked attributes, generative modeling accounts for layout logic and intuitive visual appeal by generating layouts holistically. Hence, this approach has recently gained increasing prominence \cite{layouttf2021gupta,layouttfpp2023jiang,lin2023iccv,layoutprompter2023lin,layoutnuwa2024tang}, signaling a notable shift in the modeling strategies within this domain.

Recently, Large Language Models (LLMs) \cite{zhao2023survey} have witnessed remarkable breakthroughs in the NLP field \cite{gpt42023,phi1.52023li}. They exhibit commendable prowess on various reasoning tasks, such as commonsense reasoning \cite{causalreason2011aaai,levesque2012winograd} and logistic reasoning \cite{nijkamp2023codegen,phi12023gunasekar,phi1.52023li}. Layout generation shares fundamental requirements with these tasks: translating high-level conceptual instructions into a structured output format. The exceptional reasoning skills of LLMs essentially contribute to maintaining the visual rationality and aesthetics in generated layouts. Several pioneering efforts \cite{layoutprompter2023lin,layoutnuwa2024tang} have probed the potential of LLMs in this field. LayoutPrompter \cite{layoutprompter2023lin} leverages the off-the-shelf 175B GPT3 \cite{gpt32020brown} and prompts the frozen model to awaken its innate layout-related knowledge. LayoutNUWA \cite{layoutnuwa2024tang} exploits LLaMA2-7B \cite{llama22023touvron} and CodeLLaMA-7B \cite{codellmam2023roziere} to tap into their semantic abilities, facilitating the generation of layouts.

Nonetheless, existing LLM-based methods grapple with several key limitations. First, the employed LLMs possess massive sizes (175B and 7B parameters). While the immense capacity and rich pretrained knowledge partially undergird generation quality, it exacts the cost of computational efficiency. This incurs prohibitive resource consumption for training and hampers efficient deployment within practical layout design workflows, especially in resource-intensive cases. Second, they rely on intricate HTML-based prompts to transmute layout generation into the code completion task. However, this approach suffuses layouts with redundant code symbols, such as \texttt{<html>} and \texttt{<\//html>}, which not only obscures the understanding of valid layout information but also decelerates inference due to the substantial token increment. Third, these approaches are confined to limited tasks or data domains. As the field progresses, researchers have increasingly oriented toward building task-generic \cite{layoutdm2023Inoue,LDGM2023hui,layoutprompter2023lin,layoutnuwa2024tang} or domain-generic \cite{layoutnuwa2024tang} models, as opposed to task-specific or domain-specific ones. Despite these advancements, existing layout LLMs lack comprehensiveness either in terms of task \cite{layoutnuwa2024tang} or domain \cite{layoutprompter2023lin}, yet to unleash LLM's layout reasoning prowess among more challenging and versatile scenarios. Actually, the exploration of unified layout generation predominantly revolves around masked sequence modeling with encoder-based Transformers, leaving the effectiveness of decoder models for task-generic and domain-generic generation unexplored.

The drawbacks of existing LLM-based methods inspire two key questions: (1) Can an LLM effectively unify layout generation spanning both tasks and domains, despite increased complexities? (2) Can we use a smaller LLM to strike a better performance-efficiency balance in such a more challenging unified scenario? Driven by these inspirations, we propose Layout Generation GPT (LGGPT), a generic model dedicated to unleashing the reasoning expertise of LLM in unified layout generation, based on a smaller LLM. First, we devise the Arbitrary Layout Instruction (ALI) and Universal Layout Response (ULR) as the uniform I/O template. ALI not only accommodates arbitrary layout generation inputs, thus encompassing all possible generation tasks but also spans the generation of multi-domain layouts. The collaboration of ALI and ULR enables LGGPT to generate complete layouts given any layout conditions without specific task guidance for multiple layout domains, therefore unifying both task-generic and domain-generic layout generation as a universal engine. This ventures into the widest and most challenging scenario hitherto unexplored. Second, ALI boasts a much more compact structure than prior HTML-based instructions \cite{layoutprompter2023lin,layoutnuwa2024tang}, which eschews superfluous content prevalent in HTML such as \texttt{<html>} and \texttt{<\//html>}, \texttt{<body>} and \texttt{<\//body>}. This facilitates model's instruction tuning with more condensed layout knowledge and boosts generation quality. Third, we propose an Interval Quantization Encoding (IQE) strategy, which maps each geometric value into an exclusive interval, ensuring discriminability between geometric values. This guarantees that known geometric values remain identifiable to the model, thus eliminating the need for placeholders to represent unknown values that should be predicted. Consequently, it refines the ALI to be a more succinct and informative format, significantly enhancing model’s unified layout generation performance. Furthermore, while enlarging the scale of LLM typically leads to better performance, it inevitably sacrifices model efficiency. To surmount this, we explore employing a smaller-scale LLM with 1.5B parameters, in an attempt to attain satisfactory performance while exercising better computational frugality.

In our experiments, we unify layout data from four domains, namely, scientific article, App UI, magazine, and slide, with five layout datasets, including PubLayNet \cite{publaynet2019zhong}, Rico \cite{rico2017deka}, Magazine \cite{magazine2019zheng}, SPaSe \cite{spase2019haurilet}, and WiSe \cite{wise2019haurilet}. LGGPT is trained with all tasks and all domains of data in tandem. For evaluation, we assess LGGPT under separate tasks as well as hybrid tasks, covering evaluations with arbitrary layout inputs. Experiments demonstrate that, as a generic model, LGGPT delivers superior or competitive performance to prevailing domain-specific or task-specific methods, proving that a smaller LLM can beat the previously used much larger LLMs. We reveal that IQE and ALI can significantly improve LGGPT's performance, as evidenced by comparisons with other implementation variants. Furthermore, we verify the necessity for LLMs to handle the challenging unified layout generation. Through comparisons of LLMs of various scales, we validate that a 1.5B parameter size could strike an optimal performance-efficiency trade-off in the current unified scenario.

To summarize, our main contributions include:
\begin{itemize}
	\item We propose LGGPT, a generic LLM-based model that, for the first time, achieves both task-generic and domain-generic unification layout generation.
	\item We devise the ALI and ULR as the uniform I/O template, and propose an IQE strategy to streamline layout inputs. ALI and ULR support layout generation given arbitrary layout conditions of any domain. IQE compresses ALI into a more succinct and informative structure, facilitating instruction tuning on LGGPT with condensed layout knowledge and essentially boosting generation performance.
	\item Experiments demonstrate that LGGPT yields state-of-the-art or comparable performance compared to existing methods. We successfully strike an excellent performance-efficiency trade-off with a smaller LLM, which beats much larger layout LLMs even in such a more challenging scenario.
	\item We demonstrate the necessity of exploiting LLMs to tackle the complex, varying unified layout generation. Additionally, we manifest that 1.5B parameters could be an optimal balance between model proficiency and efficiency, potentially representing a sweet spot for smaller LLMs in this scenario.
\end{itemize}

\section{Related Work}
\label{sec::related}
\subsection{Layout Generation}
Automatic layout generation has emerged as a burgeoning research topic for its extensive application in diverse scenarios, such as print publications \cite{publaynet2019zhong,magazine2019zheng,read2020patil,vtn2021arroyo}, poster/advertisement design \cite{poster2021guo,poster2021qian}, and graphic user interface design \cite{rico2017deka,ruite2021rahman,layoutganpp2021kikuchi,ganbased2021lis}. Based on generation requirements, layout generation can be broadly classified into conditional and unconditional generation. Conditional generation more specifically encompasses tasks including layout completion \cite{layouttf2021gupta,layoutnuwa2024tang}, relationship control \cite{lee2020neural,layoutganpp2021kikuchi,layoutdm2023Inoue}, and noise refinement \cite{lee2020neural,ruite2021rahman,layoutprompter2023lin}, \emph{etc.}, all of which require generating layouts predicated on specific conditions. In contrast, unconditional generation refers to crafting a new layout from scratch without any prior information. Earlier layout generation methods involve classic optimization on energy-based models \cite{convetion2014donovan,convention2015donovan}, as well as building models based on Generative Adversarial Network (GAN) \cite{layoutgan2018li,magazine2019zheng,nauata2020house,ganbased2021lis,layoutganpp2021kikuchi} and Variational AutoEncoder (VAE) \cite{layoutvae2019jyothi,lee2020neural,read2020patil}. Currently, Transformer-based approaches dominate the state-of-the-art of this field, which utilizes the self-attention mechanism \cite{attention2017vaswani} to learn the contextual relationship between layout objects and enhance generation quality. Contingent on the modeling paradigm, they can be generally grouped into masked modeling-based and generative modeling-based methods.

\textbf{Masked modeling}. In this paradigm, layout sequences undergo a masking process to create partial inputs, requiring the model to predict masked attributes and construct complete layouts. This methodology parallels the principle of masked language modeling epitomized by BERT \cite{bert2019devlin}, thus they typically employ Transformer encoder as the core model component. For example, LayoutGAN++ \cite{layoutganpp2021kikuchi} embeds Transformer into a GAN framework and performs latent optimization on the relational control task. BLT \cite{blt2022kong} discovers the immutable dependency chain problem that prevents autoregressive decoders from conditional generation and leverages a bi-directional Transformer to surmount this issue. More recently, diffusion model \cite{diffusion2020jonathan} has seen an exponential surge in the research community. A deluge of approaches have emerged to exploit this technique in concert with Transformer \cite{layoutdm2023Inoue,LDGM2023hui,layoutdiff2023zhang,layoutdm2023chai,dlt2023levi}. They corrupt the layout attributes through a Markov process to simulate different generation task conditions and perform reverse denoising from timestep $T$ until 0 to derive complete layouts. For instance, LDGM \cite{LDGM2023hui} decouples diffusion forward processes for each attribute parallelly and conducts a joint denoising process with global context to improve generation. LayoutDiffusion \cite{layoutdiff2023zhang} proposes a block-wise transition matrix based on the heterogeneous nature of layout to realize a mild forward process, thus easing the attribute estimation in the reversed process.

\textbf{Generative modeling}. Methods under the generative paradigm involve holistically generating layouts in a predict-next manner, usually exploiting Transformer of either encoder-decoder or decoder-only architectures. Corase-to-Fine \cite{ctf2022jiang} generates layout latent code with an encoder and performs corase-to-fine decoding by a two-stage decoder. LayoutFormer++ \cite{layouttfpp2023jiang} utilizes a bi-directional encoder with a decoder to perform various generation tasks. Parse-Then-Place \cite{lin2023iccv} proposes a two-stage approach to decompose text-to-layout tasks based on a T5 \cite{t52020jmlr} model. In contrast to the above encoder-decoder models, decoder-only models have received less attention until the transformative breakthrough of LLM, with only LayoutTransformer \cite{layouttf2021gupta} and VTN \cite{vtn2021arroyo} having employed this architecture. Propelled by the monumental success of LLM, LayoutPrompter \cite{layoutprompter2023lin} proposes to exploit the hidden layout cognitive ability inside the frozen GPT3 for generation tasks. LayoutNUWA \cite{layoutnuwa2024tang} utilizes LLaMA2 \cite{llama22023touvron} and CodeLLaMA \cite{codellmam2023roziere}, performing layout generation based on code instruction tuning.

\textbf{Task \& Domain Unification.} From the perspective of unification, research efforts have undergone a perceptible transition from building task-specific model \cite{lee2020neural,layouttf2021gupta,vtn2021arroyo,layoutganpp2021kikuchi,ctf2022jiang} to task-generic model \cite{blt2022kong,layoutdm2023Inoue,LDGM2023hui,layouttfpp2023jiang,layoutprompter2023lin,layoutnuwa2024tang}. BLT \cite{blt2022kong} and LayoutNUWA \cite{layoutnuwa2024tang} covers a limited range of tasks, with three and four tasks respectively. LayoutPrompter \cite{layoutprompter2023lin} extends the range to solve five tasks in a training-free manner. LayoutFormer++ \cite{layouttfpp2023jiang} and LayoutDM (Inoue et al.) \cite{layoutdm2023Inoue} address six common layout generation tasks, in which LayoutFormer++ trains the same model for each task separately, whereas LayoutDM trains the model with all tasks simultaneously. LDGM \cite{LDGM2023hui} transcends the limitation of handling six fixed tasks to handling hybrid tasks, which is the combination of various separate tasks. It unifies both separate and hybrid tasks with a joint training procedure and achieves a much broader setting. The scope of research has also expanded in domain coverage. LayoutNUWA \cite{layoutnuwa2024tang} performs joint training with layout data from all three domains (scientific article, App UI, and magazine) under a domain-generic framework, whereas prior methods simply perform generation on a single-type of layout data. However, no research has ever stepped into the broadest yet challenging repertoire of unifying multiple tasks and data domains with a joint training process. For either research or industry applications, building a task-generic as well as domain-generic generation engine indeed holds significant value. Therefore, we propose LGGPT to unify various tasks and domains of layout data, and additionally incorporate a text-to-layout task \cite{lin2023iccv} under the domain-generic setting, further extending the comprehensiveness of unification.

\begin{figure*}[t]
	\centering
	\includegraphics[width=\textwidth]{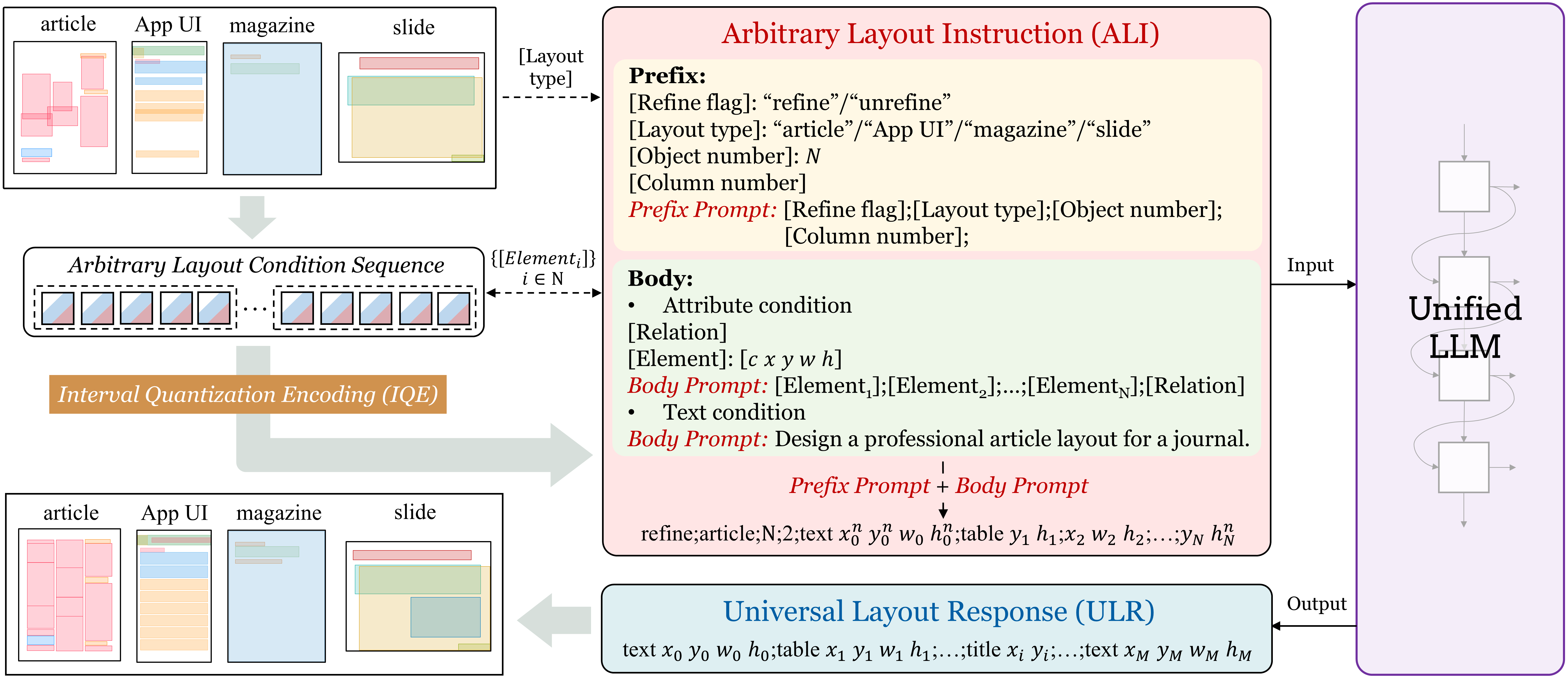}{}
	\caption{Overall architecture of LGGPT, which mainly consists of Arbitrary Layout Instruction (ALI), Universal Layout Response (ULR), the Interval Quantization Encoding (IQE) strategy, and a unified LLM. ALI is utilized for instruction tuning on the LLM, which consolidates a designated prompt for layout type and random layout conditions through \emph{Arbitrary Layout Condition Sequence}. IQE is proposed to compress ALI for a more condensed structure. ULR requires the LLM always to generate a complete, precise layout given arbitrary layout inputs.}
	\label{Fig::arch}
\end{figure*}

\subsection{Large Language Model}
\label{sec::rw_llm}
\textbf{LLM for layout reasoning.} Large Language Models (LLMs) \cite{zhao2023survey}, such as GPT4 \cite{gpt42023}, LLaMA3 \cite{llama32024dubey}, and Qwen2 \cite{qwen22024yang}, have witnessed tremendous progress in the NLP field. The success of LLMs on reasoning tasks, for example, commonsense reasoning \cite{causalreason2011aaai,levesque2012winograd} and logistic reasoning \cite{nijkamp2023codegen,phi12023gunasekar,phi1.52023li}, underscores their potential for structured reasoning more broadly. Since layout generation demands a blend of logical consistency and aesthetic sensibility, it immensely benefits from the sophisticated context understanding and causal reasoning skills of LLMs. This positions them as a compelling foundation for solving complicated layout generation tasks. Several researches have grounded the feasibility of applying LLMs in layout generation through their reasoning abilities. In natural scenarios, VISORGPT \cite{visorgpt2023xie} leverages GPT2 \cite{gpt22019radford} to learn visual layout priors of location, shape, and implicit relations between visual elements. LayoutGPT \cite{layoutgpt2023feng} injects visual commonsense into off-the-shelf ChatGPT \cite{chatgpt2022ouyang} to generate 2D and 3D layouts, then perform text-conditioned image or indoor scene generation. In document scenarios, LayoutPrompter \cite{layoutprompter2023lin} awakens the layout design ability by performing in-context learning on GPT3 with layout data. LayoutNUWA \cite{layoutnuwa2024tang} converts layout generation to code implementation to enhance semantic richness, leveraging LLaMA2 and CodeLLaMA for code instruction tuning. These endeavors inspire us to harness the exceptional reasoning skills of LLMs to tackle the challenging and yet-to-explore unified layout generation, which spans arbitrary tasks and various domains. Nevertheless, their sheer scales require substantial training resources and impede the ease of deployment. In this paper, we turn to smaller LLMs to reduce computational cost. We propose a suite of techniques to ensure respectable performance while optimizing computational efficiency, striving for a satisfactory trade-off between performance and efficiency.

\textbf{Instruction Tuning.} Instruction tuning refers to fine-tuning LLMs on a dataset of instructions and corresponding desired responses. It holds increasing significance in aligning LLMs with human preferences and has now become a key ingredient of LLMs' training recipe \cite{zhang2023instruction}. Conventionally, most LLMs perform instruction tuning on homogeneous data, \emph{i.e.}, fine-tuning the model with data reflective of their pre-training exposure, such as natural language text \cite{peng2023instruction,1_8ktasks2022chung} and code \cite{codellmam2023roziere,phi12023gunasekar}. However, a growing body of research is exploring the potential of heterogeneous data, \emph{i.e.}, data that differs significantly from pre-training content in structure, such as time-series data \cite{llmtimeseries2023hao,yu2023temporal}, to instruct LLMs for domain-specific comprehension and broader task generalization. This further extends the border of instruction tuning and enables more flexible specialties of LLMs. Converse to prior layout LLMs \cite{layoutprompter2023lin,layoutnuwa2024tang} that adapt layout sequences into the HTML code format replete with superfluous code symbols, we formulate layouts into a succinct composition of solely textual classes and numerical geometries. We then encompass them within our ALI and ULR templates for more compact and effective layout representations. Although this is heterogeneous data to mainstream LLMs, their innate reasoning capability learned from pretraining still benefits the understanding of layout context. Hence, building upon the success of trailblazing works \cite{llmtimeseries2023hao,yu2023temporal}, we employ heterogeneous instruction tuning to empower LLMs to understand specialized layout conditions and perform diverse layout generation tasks seamlessly.

\section{Methodology}
The overall architecture of the proposed LGGPT is illustrated in Fig.~\ref{Fig::arch}. Concretely, we propose the Arbitrary Layout Instruction (ALI) and Universal Layout Response (ULR) as the uniform I/O template, tailored to unify arbitrary layout generation tasks and multiple domains. Through instruction tuning based on ALI and ULR, LGGPT is empowered to generate complete and precise layouts of desired domains given any layout condition inputs. Additionally, we propose an Interval Quantization Encoding (IQE) strategy. It compresses ALI into a compact yet information-dense structure by preserving valid layout clues and eliminating redundant placeholders, essentially facilitating model's understanding of variable layout conditions.

\begin{figure}[t]
	\centering
	\includegraphics[width=\linewidth]{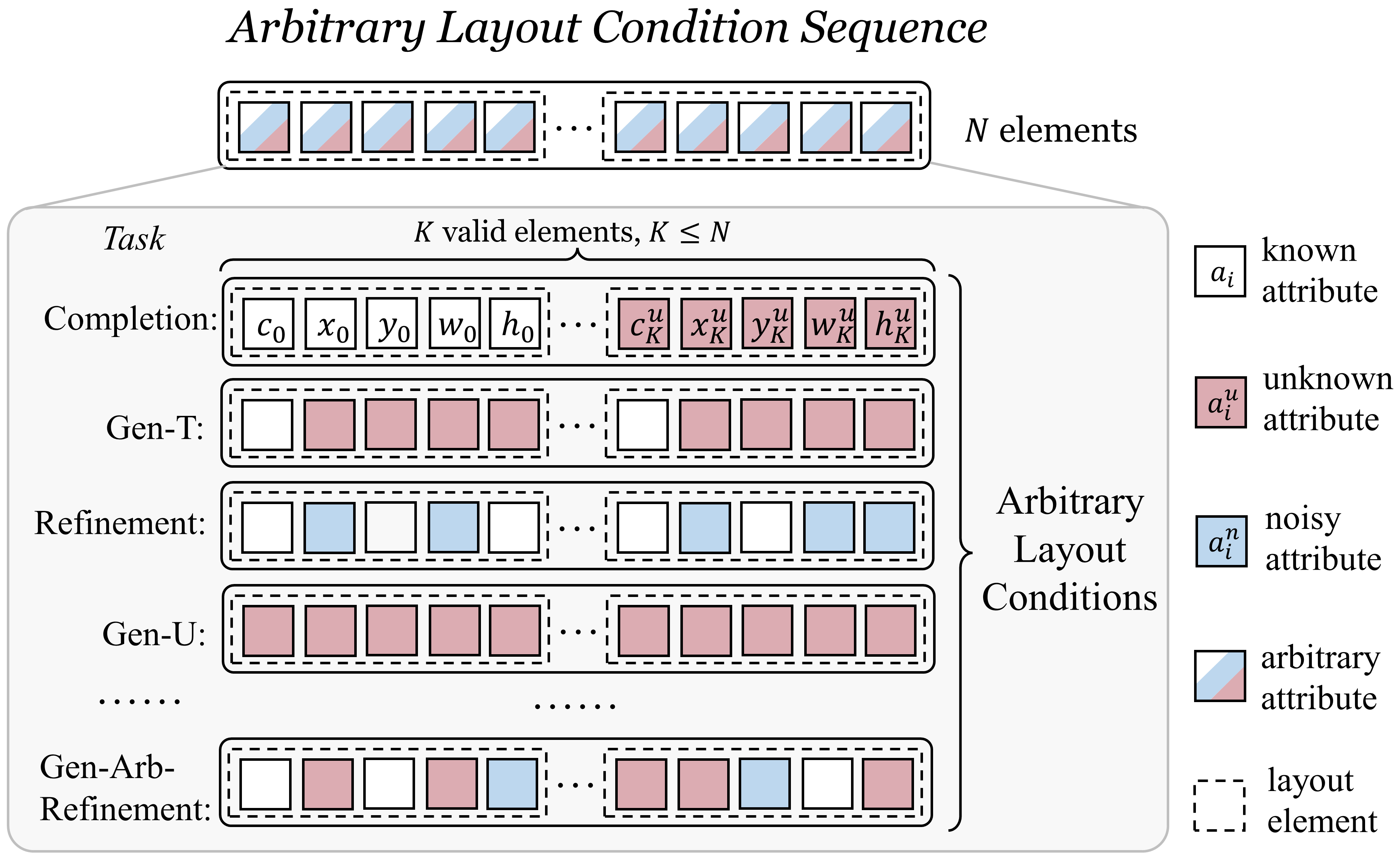}{}
	\caption{The visualized demonstration of the Arbitrary Layout Condition Sequence, which is the key component of ALI that accounts for the ``arbitrary" property. It accommodates arbitrary layout conditions by supporting the infinite combination of \emph{known}, \emph{unknown}, and \emph{noisy} attributes, therefore covering all possible layout generation tasks.}
	\label{Fig::alcs}
\end{figure}

\subsection{Layout Representation}
Generally, a layout $L$ is composed of $N$ elements, with each element characterized by five attributes, \emph{i.e.}, element class $c$, the left and top bounding box coordinates $x$ and $y$, the width $w$ and height $h$. Therefore, by flattenedly splicing $N$ elements, a layout is represented as $L=\{c_1,x_1,y_1,w_1,h_1;~...;~c_N,x_N,y_N,w_N,h_N \}$. $c$ is a textual attribute. $x$, $y$, $w$, and $h$ are usually quantized from float values to integers for model learning in the discrete space \cite{vtn2021arroyo,layouttf2021gupta}.

\subsection{Arbitrary Layout Instruction}
\label{sec::ali}
A universal system is expected to accommodate a wide range of inputs. In layout generation, conditional generation tasks showcase substantial variations among inputs due to the versatile nature of user requirements. Also, generating different domains of layout requires specific prompts to instruct the model in synthesizing intended layouts. This underscores the need for a uniform input format that comprehensively covers task and domain specifications. Therefore, we devise the Arbitrary Layout Instruction (ALI) to unite arbitrary layout inputs. As depicted in Fig.~\ref{Fig::arch}, ALI consists of the \emph{Prefix Prompt} and the \emph{Body Prompt}, in which the \emph{Body Prompt} contains the main layout information of ALI.

In the \emph{Prefix} part, \textbf{[Refine flag]} signifies whether the model should undertake the refinement task \cite{lee2020neural,ruite2021rahman}, which should be either ``refine" or ``unrefine". \textbf{[Layout type]} specifies the desired type of layout to be generated. This prompt should be one of ``article", ``App UI", ``magazine", and ``slide". Note that we use ``article" to represent the scientific article layout. \textbf{[Object number]} represents the number of elements. \textbf{[Column number]} indicates the number of layout columns, particularly relevant for layouts such as articles that typically feature multi-column structures. 

The \emph{Body} part includes attribute conditions and text conditions. Within attribute conditions, \textbf{[Relation]} represents the pairwise element relationships. \textbf{Element} denotes an element composed of the class $c$ and bounding box $x,y,w,h$. We define input layouts sequence \{\textbf{[Element$_i$]}\}, $i \in N$ as the Arbitrary Layout Condition Sequence in Fig.~\ref{Fig::arch}. This sequence embodies the essential ``arbitrary" characteristic of ALI and is further illustrated in Fig.~\ref{Fig::alcs}. Each attribute among an element is assigned one of three statuses: \emph{known}, \emph{unknown}, or \emph{noisy}. \emph{known} implies that this attribute is accurately provided. \emph{unknown} represents the absence of this attribute in the inputs. \emph{noisy} indicates that this attribute has been perturbed with noise, applicable only to $x_i$, $y_i$, $w_i$, or $h_i$, $i \in N$. The noisy attributes are denoted with superscript $no$, for instance, $x^{no}_i$. With noise added, the model is tasked with refining the attribute to its accurate value, \emph{i.e.}, eliminating the noise, and the \textbf{[Refine flag]} is accordingly set to ``refine". Attribute statuses inside this sequence could be arbitrarily designated, thus allowing any customized layout conditions as inputs and covering all possible layout generation tasks. We then concatenate all the \emph{known} and \emph{noisy} attributes and skip the \emph{unknown} ones to form the sequence, using space `` " as the separator. The concatenation is in line with the Interval Quantization Encoding strategy, which will be elaborated in Sec.~\ref{sec::iqe}. Each element is concatenated using the semicolon ``;" as the separator to construct the \emph{Body Prompt}. A special case is the unconditional generation, whose attributes are all \emph{unknown} and the \emph{Body Prompt} will be empty. For text conditions, we design several domain-specific natural language prompts, which are detailed in Appendix~\ref{sec::nlp_example}. The \emph{Prefix Prompt} and the \emph{Body Prompt} are finally merged to construct the ALI, serving as the input for instruction tuning on the unified LLM.

The elaborate structure of ALI provides sufficient inclusivity and flexibility for diverse layout generation requirements: (1) \textbf{Boundless generation potential}. ALI's Body Prompt incorporates arbitrary layout conditions, which allows ALI to cover any conceivable layout generation requirement and thus support any generation task (not limited to the 11 tasks selected for comparison in experiments (Sec.~\ref{exp::evaluate})), providing unparalleled flexibility. (2) \textbf{Task-generic intelligence}. The adaptation to unlimited layout inputs of ALI enables the model to automatically infer task types, eliminating the conventional need of using specific prompts to specify the desired task \cite{layoutprompter2023lin,layoutnuwa2024tang}, showcasing a higher degree of intelligence and versatility. (3) \textbf{Seamless domain adaptation}. The designated prompt for layout types allows for seamless adaptation across different layout domains. Equipped with ALI, LGGPT stands as the first attempt that unifies both task-generic and domain-generic layout generation, marking the broadest unification achieved in layout generation hitherto.

\begin{figure}[t]
	\centering
	\includegraphics[width=\linewidth]{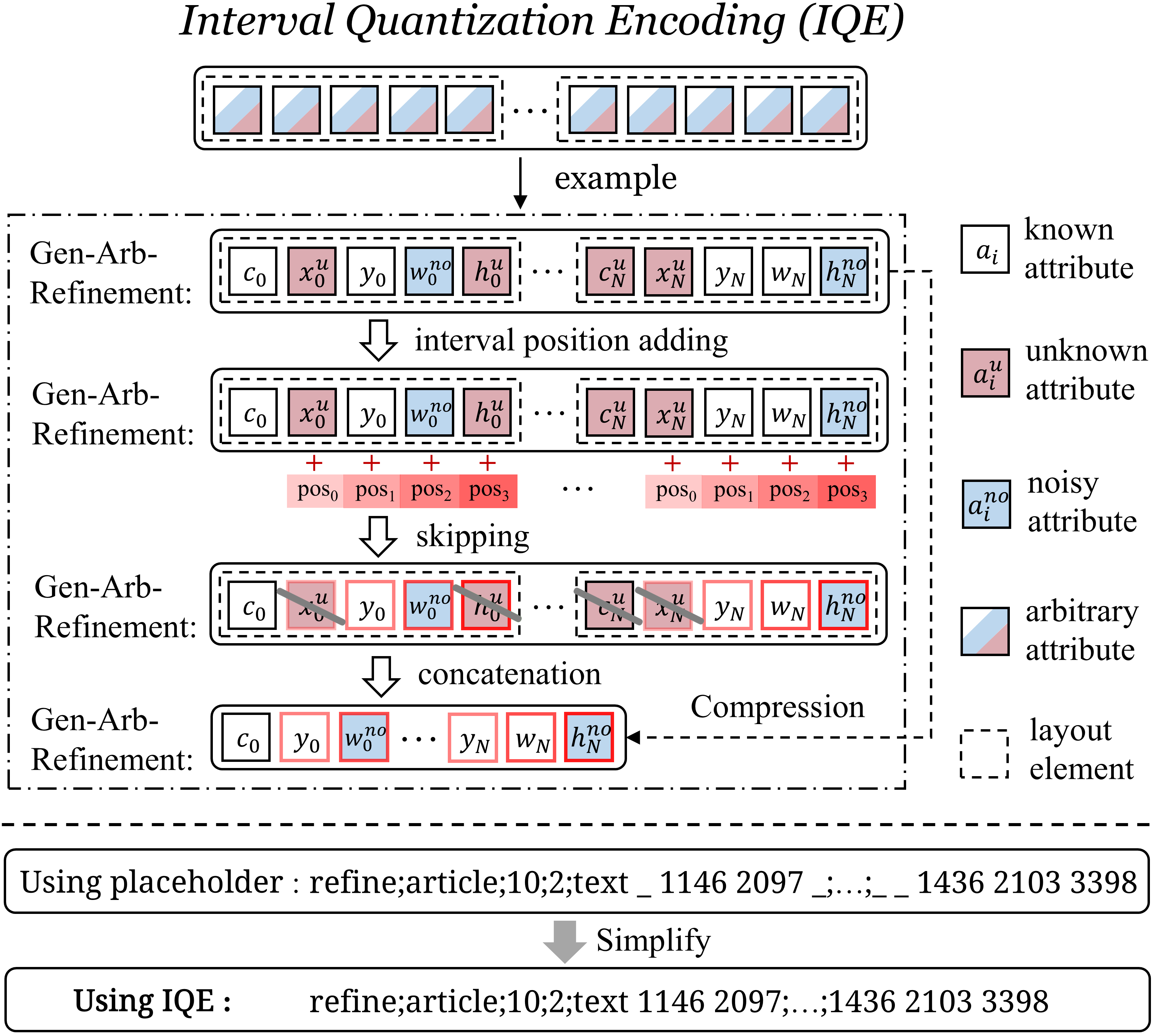}{}
	\caption{The schematic of Interval Quantization Encoding. It applies positional encoding to the $x,y,w,h$ attributes of each layout element by adding independent interval values. This enables the model to distinguish them solely according to the numerical magnitudes. Then we can skip the \emph{unknown} attributes and concatenate other attributes to compress the layout sequence, avoiding the usage of conventional placeholders and significantly increasing the valid information density of input instructions. We present an example comparison of using placeholder and IQE at the bottom.}
	\label{Fig::iqe}
\end{figure}

\subsection{Interval Quantization Encoding}
\label{sec::iqe}
Among each element, the class $c$ is represented textually, while $x,y,w,h$ are quantized integers. In conditional generation, certain attributes will be in the \emph{unknown} status, for example, $x,y$ are unknown in the generation conditioned on types and sizes (Gen-TS, will be explained in Sec.~\ref{exp::evaluate}). Directly concatenating the bounding box integers in the instruction (such as in LayoutFormer++ \cite{layouttfpp2023jiang}) could cause ambiguity, as the model might not discern which attribute corresponds to $x$ or $w$. A typical solution is placing placeholders on the positions of unknown content as indicators. However, it significantly introduces superfluous symbols in layout representation that hampers layout knowledge learning of the model, especially when a large portion of attributes is unknown. To address this, we propose an Interval Quantization Encoding (IQE) strategy to bypass introducing redundant placeholders while retaining all valid layout attributes in ALI. Fig.~\ref{Fig::iqe} shows the schematic of IQE. We first compute the maximum side length $l_m$ of all layout pages. Then we scale the values of $x,y,w,h$ by adding $l_m$ during quantization with the following rule:
\begin{equation}
	\label{eq::iqe}
	\begin{aligned}
		&pos_{i} = i \cdot l_m, i \in \{0,1,2,3\}\\
		&\hat{x} = x + pos_0;~\hat{y} = y + pos_1;\\
		&\hat{w} = w + pos_2;~\hat{h} = h + pos_3.
	\end{aligned}
\end{equation}
\noindent $pos_i$ serves as a positional encoding for $\hat{x},\hat{y},\hat{h},\hat{w}$, ensuring that each of them resides within an exclusive and independent interval $[i \cdot l_m, (i + 1) \cdot l_m)~i \in \{0,1,2,3\}$. This arrangement carves out distinct intervals for each attribute, allowing them to be uniquely identified solely based on their numerical values. Hence, we can omit the \emph{unknown} attributes in the prompts, \emph{i.e.}, using the absence of information to represent the \emph{unknown} values instead of resorting to explicit placeholders like ``\_" or ``-". The \emph{known} and \emph{noisy} attributes can be directly concatenated without confusing the model. For example, if we use the placeholder ``\_" to represent \emph{unknown} attributes, an input prompt before tokenization could be ``refine;article;10;2;text \_ 1146 2097 \_;…;\_ \_ 1436 2103 3398". Here, figures like 1046 and 2097 are the known information (the $y$ coordinate and width here) of a layout element, and the placeholders are placed on the positions of unknown ones. Upon the utilization of IQE, the input prompt is simplified to ``refine;article;10;2;text 1146 2097;…;1436 2103 3398", where the unknown information is excluded in the prompt.

As shown in Fig.~\ref{Fig::arch}, IQE is utilized as an inter-connected component of ALI to streamline its structure, enhancing the richness of layout-relevant information within the prompt by removing placeholders and benefiting the model to grab valid layout knowledge. Compared with the HTML code format used in \cite{layoutprompter2023lin,layoutnuwa2024tang}, ALI already forgoes the excess code syntax such as \texttt{<html>} and \texttt{<\//html>}, \texttt{<body>} and \texttt{<\//body>} in its basic format. The incorporation of IQE further compresses the length and refrains from the content-sparse placeholders in the \emph{Body Prompt}, bestowing upon ALI a far more succinct and concentrated structure. This simplification significantly increases the information density in layout prompts, bolstering the model's ability to understand complex and variable layout generation conditions during the unified training process.

\subsection{Universal Layout Response}
\label{sec::ulr}
We introduce the Universal Layout Response (ULR) template for a uniform generation output format, as depicted at the bottom of Fig.~\ref{Fig::arch}. Regardless of the input layout conditions, ULR supervises the model to always output complete layouts during both training and inference. The \emph{unknown} attributes are required to be predicted and the \emph{noisy} attributes should be denoised to precise ones. The outputted element classes should fall into the category group of the layout domain intended to be generated. Therefore, ULR adapts to any layout generation task of interest and handles the synthesis for any domain of layouts.

During training, we use teacher forcing by joining the ALI and ULR using the separator \#. Here, ULR is the ground truth of the target layout. To align with ALI, IQE is also applied in ULR. Therefore, during inference, the ULR is encoded with the interval positions (Eq.~\ref{eq::iqe}) and we need to subtract the $pos_{i} = i~\cdot~l_m, i \in \{0,1,2,3\}$ from the geometric attributes of each element to revert to the native predicted geometries for metric computation or layout rendering. Significantly, ALI together with ULR establish the uniform I/O format for LGGPT. ULR can be autoregressively predicted after the ALI, thus fundamentally addressing the immutable chain dependency problem of autoregressive models \cite{blt2022kong} and providing a well-suited I/O representation for decoder-only LLMs.

\subsection{Model Training}
\subsubsection{Architecture}
In recent years, there has been a remarkable surge in Large Language Models (LLMs), including the GPT family \cite{gpt12018radford,gpt22019radford,gpt32020brown,gpt42023}, PaLM family \cite{palm2023akanksha,palm22023anil}, LLaMA family \cite{llama2023touvron,llama22023touvron}, \emph{etc.} Despite most LLMs possessing over 7B parameters, a growing body of smaller LLMs with 1B-3B parameters have been progressively proposed. These smaller models represent a strategic balance between cognitive depth and computational demand. To explore a suitable balance between model proficiency and computational efficiency, we opt for GPT2-XL \cite{gpt22019radford}, a more compact LLM with 1.5B parameters, as the core model of LGGPT. We leverage its pretrained weights to enhance generalization performance by leveraging its pre-existing general understanding capabilities.

\subsubsection{Training and Optimization}
As illustrated in Sec.~\ref{sec::ulr}, we employ teacher forcing by appending the ground truth of outputs to the instruction prompts to form model inputs. We solely optimize the probability of predicted tokens converging toward the ground truth layout tokens, and omit the optimization of the prompt part, \emph{i.e.}, the tokens preceding the separator token \#. The model is optimized by minimizing the negative log-likelihood of the predicted layout tokens $t$:
\begin{equation}
	\mathcal{L}=-\sum_k^K logP(t_k | t_{1:k-1};\Theta), k \in K
\end{equation}
where $K$ denotes the length of the output token sequence, $p(\cdot | \cdot)$ represents the conditional probability, and $p(t_k | t_{1:k-1})$ indicates the probability of the current token decided by all previous tokens. $\Theta$ denotes the model parameters.

\subsection{Decoding Scheme}
A variety of decoding strategies have been developed to enhance performance in autoregressive generation tasks, such as greedy search, beam search \cite{beam1994steinbisss}, multinomial sampling \cite{multinomial1968saul}, Top-k sampling \cite{topk2018fan}, and Top-p sampling \cite{topp2020ari}. In our experiments, unless otherwise specified, we default to using greedy search as the basic decoding scheme and employ Top-k sampling with $k=50$. When invoking Top-k sampling, the temperature of softmax \cite{attention2017vaswani} is set to $1.0$.

\section{Experiment}
\label{sec::exp}
\subsection{Dataset}
We conduct experiments using five public datasets: PubLayNet \cite{publaynet2019zhong}, Rico \cite{rico2017deka}, Magazine \cite{magazine2019zheng}, SPaSe \cite{spase2019haurilet}, and WiSe \cite{wise2019haurilet}. PubLayNet \cite{publaynet2019zhong} contains 346K+ annotated scientific article pages of five element types. Rico \cite{rico2017deka} contains 66K+ graphics user interfaces for mobile applications (App UI) of 25 element types. Magazine \cite{magazine2019zheng} consists of 4K magazine pages of six element types. SPaSe \cite{spase2019haurilet} and WiSe \cite{wise2019haurilet} both consist of presentation slides, with 2,000 and 1,300 images of the same 24 element types, respectively.

Similar to prevailing works \cite{LDGM2023hui,layoutdm2023Inoue,blt2022kong}, we perform pre-filtering on these datasets and then split them into the training and testing sets. For PubLayNet, following \cite{layoutdm2023Inoue,layoutnuwa2024tang}, we filter out samples with more than 25 elements and utilize the entire official ``train" set of PubLayNet for training while using the ``val" set for testing. The Rico data has no official data splitting. We filter out samples with more than 40 elements and follow most existing works \cite{layoutganpp2021kikuchi,layoutdm2023Inoue,layoutnuwa2024tang} to assign 90\% data for training while using the rest 10\% for testing. For Magazine, we remove the \emph{background} element category and discard samples with more than 24 elements. Since it has no official splitting either, we similarly split it at a ratio of 9:1. We consolidate the SPaSe and WiSe datasets as a cohesive set composed of slide data, and denote it as the Slide dataset, which has 3,329 slide images of 24 element classes. The Slide dataset is also split at the ratio of 9:1. The consistent data splitting ratio ensures the same testing data for fair comparisons. Ultimately, the splits yield 333,848/11,208 samples for PubLayNet, 47,028/5,226 for Rico, 3,524/392 for Magazine, and 3,299/333 for Slide, respectively.

\subsection{Evaluation Task}
\label{exp::evaluate}
We perform assessments on distinct layout datasets, by specifying the \textbf{[Layout type]} description in ALI to match the corresponding domain of layout. We adhere to the prescribed settings of domain-specific models to evaluate six separate tasks \cite{layoutdm2023Inoue,layouttfpp2023jiang,layoutprompter2023lin}. Additionally, we explore hybrid tasks \cite{LDGM2023hui}, which combine a spectrum of separate tasks into a more general setting.

\begin{figure}[t]
	\centering
	\includegraphics[width=\linewidth]{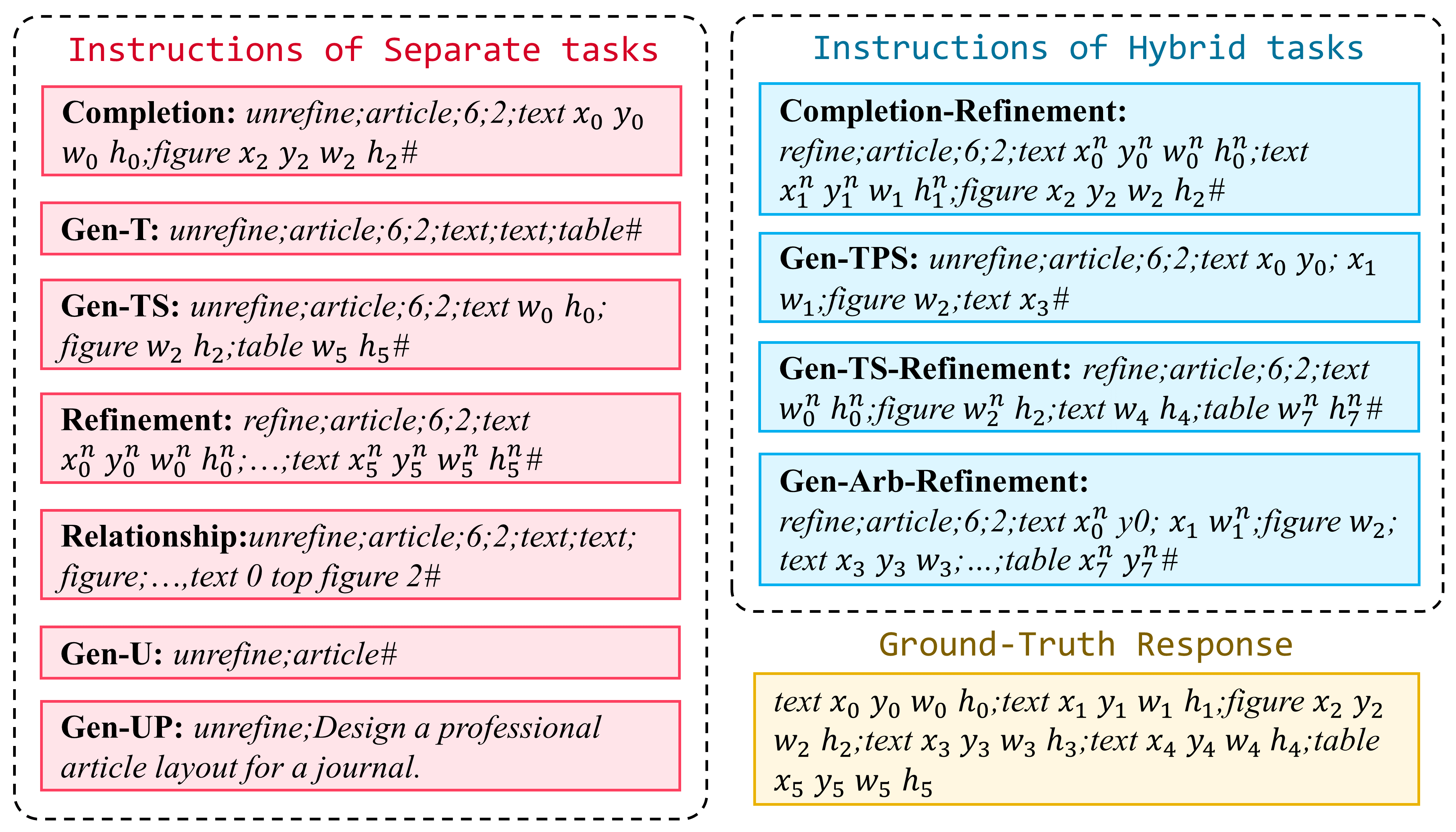}{}
	\caption{A detailed demonstration of the evaluated tasks and the corresponding instructions. $x^{no}_i$ denotes the noisy attribute. The ground-truth response represents the complete layout.}
	\label{Fig::task_prompt}
\end{figure}

\noindent \textbf{Separate tasks:}
\begin{itemize}
	\item \textbf{Completion} is generating layouts given partial elements with all known attributes. For a complete layout, we randomly sample a portion of all elements as input, treating the remaining elements as the target to be generated.
	\item \textbf{Gen-T} is generating layouts conditioned on the given classes of arbitrary elements.
	\item \textbf{Gen-TS} is generating layouts based on the given classes and sizes of arbitrary elements.
	\item \textbf{Relation} refers to generating layouts conditioned on the class of each element and pairwise relation constraints. Similar to LayoutGAN++ \cite{layoutganpp2021kikuchi}, we adopt the location (top, bottom, left, right, overlapped) and size (smaller, larger, equal) relations. To simulate real-world scenarios, we randomly sample two pairs of relations for experiments.
	\item \textbf{Refinement} is conditioned on a noisy layout whose geometric information is perturbed. The model is asked to denoise and generate a fine one. Following RUITE \cite{ruite2021rahman}, we add random noise sampled from a standard normal distribution (mean: 0, standard deviation: 0.01) to the positions and sizes of layouts. We set the \textbf{[Refine flag]} in prompts to ``refine" here but ``unrefine" in other tasks. We use the multinomial decoding \cite{multinomial1968saul} without Top-k sampling for this task exclusively.
	\item \textbf{Gen-U} is generating layouts without any layout attribute constraint. Within the prompt, we solely retain \textbf{[Layout type]} to designate the intended layout domain for generation, while omitting the \textbf{[Object number]} and \textbf{[Column number]} constraints.
	\item \textbf{Gen-UP} is performing Gen-U with natural language prompts, which could be viewed as a simplified form of the text-to-layout generation task. We devise some natural language prompts, \emph{e.g.} \emph{Design a flexible layout for a magazine publisher.}, as model inputs to perform unconditional generation. Detailed text prompts are included in Appendix~\ref{sec::nlp_example}.
\end{itemize}
\noindent Some previous studies \cite{blt2022kong,layoutdm2023Inoue} have input all the classes (and sizes) of elements in Gen-T and Gen-TS. However, we adopt a more challenging setting by inputting a random number of classes (and sizes) rather than giving all of them, which we think is a more general setup.

\noindent \textbf{Hybrid tasks:}
\begin{itemize}
	\item \textbf{Completion-Refinement} is the combination of the Completion and Refinement tasks.
	\item \textbf{Gen-TPS} is generating layouts conditioned on arbitrary \emph{known} attributes (classes, positions, sizes) of random elements, without any noise added.
	\item \textbf{Gen-PS-Refinement} is generating layouts conditioned on positions and sizes of arbitrary elements that are randomly perturbed with noise, requiring the model to generate denoised and complete layouts. Note that part of the positions and sizes are in the \emph{noisy} status.
	\item \textbf{Gen-Arb-Refinemnt} is the combination of Gen-TPS and Refinement. Element attributes are arbitrarily given or perturbed. It differs from Refinement in adding geometric noise with a probability of 0.5 rather than always adding noise for a more general setting.
\end{itemize}
\noindent Although LDGM \cite{LDGM2023hui} has introduced three statuses: precise (P), coarse (C), and missing (M) to delineate the hybrid scenarios, the definitions of hybrid tasks remain ambiguous. Compared to \cite{LDGM2023hui}, we optimize the definitions of the hybrid tasks for better clarity with the exact combination of commonly used separate tasks. The  Gen-Arb-Refinement is the most comprehensive scenario that encapsulates arbitrary layout input conditions and thus all separate tasks, similar to the Gen-PCM configuration in LDGM. The hybrid task setting closely reflects the ``arbitrary" property intrinsic to our proposed ALI in task unification, serving as a robust measure of LGGPT's ability in the unified and general scenario. A more illustrative description of the tasks and corresponding input instructions is shown in Fig.~\ref{Fig::task_prompt}.

\subsection{Evaluation Metric}
\label{exp::metric}
We adopt four commonly used metrics in evaluation. 
\begin{itemize}
	\item \emph{Fréchet Inception Distance (FID)} \cite{fid2017heusel} gauges the distributional similarity between generated layouts and their corresponding ground truth counterparts in a high-dimensional feature space. We follow \cite{layoutdm2023Inoue} and use an improved FID computation method \cite{layoutganpp2021kikuchi}.
	\item \emph{Alignment} \cite{layoutganpp2021kikuchi} computes an alignment degree to measure how well the elements are aligned by the center or edge.
	\item \emph{Overlap} \cite{layoutganpp2021kikuchi} is used to measure the average overlapped degree between all pairs of bounding boxes in the layouts.
	\item \emph{Maximum IOU (Max~IOU)} \cite{layoutganpp2021kikuchi} computes the similarity of element bounding boxes with generated layouts and their ground truths. It finds the optimal matching between two layouts to compute the intersection of union, only considering elements with the identical label set. A valid comparison necessitates an exact match of both categories and the sequencing of elements between the output and the ground truth (GT). During inference, since the testing input derives from GT, Max IOU can effectively quantify the consistency between input and output.
\end{itemize}
\noindent FID, Alignment, and Overlap are the lower the better, while Max IOU is the opposite. For fair comparisons, we adopt the code implementation from \cite{layoutganpp2021kikuchi} of these metrics, which is also used in \cite{layoutdm2023Inoue}. Note that the absolute values of Alignment and Overlap are scaled by 100$\times$ for better visibility. In experiments, for the Alignment metric in some works that are not normalized, we normalize it to ensure that all metrics are on a comparable scale. Similarly, for the Overlap metric that has not been scaled by 100$\times$, we re-scale it by 100$\times$.

\begin{table*}[ht]
	\renewcommand{\arraystretch}{1.1}
	\caption{Comparison of LGGPT with state-of-the-art methods on isolated datasets. Align. denotes the Alignment metric. R. Score denotes the Ranking Score for a more intuitive ranking demonstration. $\downarrow$ signifies that smaller values are better, whereas $\uparrow$ represents the contrast. Arch denotes Transformer architecture. \textcolor{red1}{Enc}, \textcolor{yellow1}{E-D}, and \textcolor{blue1}{Dec} denote the encoder-only, encoder-decoder, and decoder-only architectures, respectively. T-G denotes ``Task-Generic" and D-G denotes ``Domain-Generic", representing different training paradigms. The best results are marked in \textbf{bold} and the second-best results are marked with \underline{underline}.}
	\label{Table::isolated_comp}
	\centering
	\resizebox*{\linewidth}{!}{
	\begin{tabular}{c c c c c c c c c c >{\columncolor[RGB]{240,240,240}}c c c c >{\columncolor[RGB]{240,240,240}}c c c c c >{\columncolor[RGB]{240,240,240}}c}
		\toprule
		\multirow{2.4}{*}{Task} & \multirow{2.4}{*}{Method} & \multirow{2.4}{*}{Venue} & \multirow{2.4}{*}{Arch} & \multirow{2.4}{*}{T-G} & \multirow{2.4}{*}{D-G} & \multicolumn{5}{c}{PubLayNet \cite{publaynet2019zhong}} & \multicolumn{4}{c}{Rico \cite{rico2017deka}} & \multicolumn{5}{c}{Magazine \cite{magazine2019zheng}}\\
		\cmidrule(r){7-11}\cmidrule(r){12-15}\cmidrule{16-20}
		~ & ~ & ~ & ~ & ~ & ~ & FID $\downarrow$ & Align. $\downarrow$ & Overlap $\downarrow$ & Max IOU $\uparrow$ & R. Score $\downarrow$ & FID $\downarrow$ & Align. $\downarrow$ & Max IOU $\uparrow$ & R. Score $\downarrow$ & FID $\downarrow$ & Align. $\downarrow$ & Overlap $\downarrow$ & Max IOU $\uparrow$ & R. Score $\downarrow$\\
		\hline
		\scriptsize \emph{Single tasks}\\
		\hline
		\multirow{9}{*}{Completion} & LayoutTransformer \cite{layouttf2021gupta} & ICCV'21 & \textcolor{blue1}{Dec} & $\usym{2718}$ & $\usym{2718}$ & 8.36 & - & - & 0.45 & 6.50 & \underline{3.71} & - & 0.54 & 5.00 & - & - & - & - & -\\
		~ & BLT \cite{blt2022kong} & ECCV'22 & \textcolor{red1}{Enc} & $\usym{2718}$ & $\usym{2718}$ & 131.00 & - & - & 0.35 & 9.00 & 117.00 & - & 0.47 & 9.00 & - & - & - & - & -\\
		~ & LayoutFormer++ \cite{layouttfpp2023jiang} & CVPR'23 & \textcolor{yellow1}{E-D} & $\usym{2718}$ & $\usym{2718}$ & 10.25 & 0.29 & \textbf{0.22} & 0.47 & 4.00 & 4.57 & 1.10 & \underline{0.73} & \underline{3.33} & - & - & - & - & -\\
		~ & LayoutNUWA-DS \cite{layoutnuwa2024tang} & ICLR'24 & \textcolor{blue1}{Dec} & $\usym{2718}$ & $\usym{2718}$ & 7.23 & 0.17 & 13.11 & 0.47 & 4.25 & 8.73 & 0.01 & 0.64 & 3.67 & 7.34 & - & - & 0.41 & 3.00\\
		\cline{5-20}
		~ & LayoutDM \cite{layoutdm2023Inoue} & CVPR'23 & \textcolor{red1}{Enc} & $\usym{2713}$ & $\usym{2718}$ & 7.65 & - & - & 0.38 & 6.50 & 9.00 & - & 0.58 & 7.00 & - & - & - & - & -\\
		~ & LDGM \cite{LDGM2023hui} & CVPR'23 & \textcolor{red1}{Enc} & $\usym{2713}$ & $\usym{2718}$ & 25.31 & \underline{0.10} & 19.45 & 0.44 & 5.75 & 16.42 & 0.36 & 0.60 & 6.00 & 24.35 & \underline{0.49} & {39.26} & 0.38 & 3.00\\
		~ & LayoutPrompter \cite{layoutprompter2023lin} & NeurIPS'23 & \textcolor{blue1}{Dec} & $\usym{2713}$ & $\usym{2718}$ & \underline{2.13} & 0.33 & \underline{1.70} & \underline{0.48} & 3.00 & 7.32 & 1.18 & 0.67 & 4.33 & - & - & - & - & -\\
		~ & LayoutNUWA-DA \cite{layoutnuwa2024tang} & ICLR'24 & \textcolor{blue1}{Dec} & $\usym{2718}$ & $\usym{2713}$ & 6.93 & 0.13 & 12.92 & \underline{0.48} & \underline{3.00} & 7.54 & \textbf{0.10} & 0.62 & 4.00 & \textbf{7.57} & - & - & \textbf{0.50} & \textbf{1.00}\\
		\rowcolor{lightblue}
		~ & LGGPT (Ours) & This Work & \textcolor{blue1}{Dec} & $\usym{2713}$ & $\usym{2713}$ & \textbf{2.08} & \textbf{0.04} & 5.54 & \textbf{0.57} & \cellcolor{colred}\textbf{1.50} & \textbf{1.03} & \underline{0.12} & \textbf{0.80} & \cellcolor{colred}\textbf{1.67} & \underline{8.11} & \textbf{0.44} & \textbf{30.60} & \underline{0.47} & \cellcolor{colred}\underline{1.50}\\
		\hline
		
		\multirow{8}{*}{Gen-T} & LayoutFormer++ \cite{layouttfpp2023jiang} & CVPR'23 & \textcolor{yellow1}{E-D} & $\usym{2718}$ & $\usym{2718}$ & 8.41 & 0.29 & \underline{0.80} & 0.35 & 5.50 & \textbf{1.10} & 3.24 & 0.43 & 4.33 & - & - & - & - & -\\
		~ & LayoutDiffusion \cite{layoutdiff2023zhang} & ICCV'23 & \textcolor{red1}{Enc} & $\usym{2718}$ & $\usym{2718}$ & \underline{3.73} & \underline{0.03} & \textbf{0.50} & 0.34 & \textbf{3.00} & \underline{1.56} & 0.12 & 0.35 & 4.00 & - & - & - & - & -\\
		~ & LayoutNUWA-DS \cite{layoutnuwa2024tang} & ICLR'24 & \textcolor{blue1}{Dec} & $\usym{2718}$ & $\usym{2718}$ & 6.72 & 0.12 & 5.30 & 0.39 & 4.25 & 3.71 & 0.12 & 0.38 & 5.33 & \underline{8.99} & - & - & 0.29 & 2.50\\
		\cline{5-20}
		~ & LayoutDM \cite{layoutdm2023Inoue} & CVPR'23 & \textcolor{red1}{Enc} & $\usym{2713}$ & $\usym{2718}$ & 7.95 & 0.12 & 18.33 & 0.31 & 6.50 & 3.55 & 0.18 & 0.28 & 6.33 & - & - & - & - & -\\
		~ & LDGM \cite{LDGM2023hui} & CVPR'23 & \textcolor{red1}{Enc} & $\usym{2713}$ & $\usym{2718}$ & 20.69 & 0.15 & 16.88 & \textbf{0.44} & 5.50 & 16.64 & 0.39 & \underline{0.58} & 5.33 & 24.67 & \textbf{0.45} & 45.11 & \textbf{0.36} & 2.00\\
		~ & LayoutPrompter \cite{layoutprompter2023lin} & NeurIPS'23 & \textcolor{blue1}{Dec} & $\usym{2713}$ & $\usym{2718}$ & \textbf{3.02} & 0.53 & 4.70 & 0.38 & 4.25 & 3.23 & 1.56 & 0.43 & 5.33 & - & - & - & - & -\\
		~ & LayoutNUWA-DA \cite{layoutnuwa2024tang} & ICLR'24 & \textcolor{blue1}{Dec} & $\usym{2718}$ & $\usym{2718}$ & 6.58 & \textbf{0.01} & 8.60 & 0.39 & \underline{3.50} & 2.52 & \textbf{0.06} & 0.45 & \underline{2.67} & \textbf{8.79} & - & - & \underline{0.31} & \textbf{1.50}\\
		\rowcolor{lightblue}
		~ & LGGPT (Ours) & This Work & \textcolor{blue1}{Dec} & $\usym{2713}$ & $\usym{2713}$ & 5.94 & 0.06 & 5.20 & \underline{0.41} & \cellcolor{colred}\textbf{3.00} & 2.45 & \underline{0.10} & \textbf{0.61} & \cellcolor{colred}\textbf{2.00} & {9.33} & \underline{0.46} & \textbf{31.58} & \textbf{0.36} & \cellcolor{colred}\underline{1.75}\\
		\hline
		
		\multirow{9}{*}{Gen-TS} & LayoutTransformer \cite{layouttf2021gupta} & ICCV'21 & \textcolor{blue1}{Dec} & $\usym{2718}$ & $\usym{2718}$ & 16.90 & 0.11 & 22.00 & 0.32 & 7.50 & 3.73 & 0.20 & 0.32 & 7.00 & - & - & - & - & -\\
		~ & BLT \cite{blt2022kong} & ECCV'22 & \textcolor{red1}{Enc} & $\usym{2718}$ & $\usym{2718}$ & 5.10 & \underline{0.08} & 19.9 & 0.39 & 5.50 & 4.48 & 0.21 & 0.34 & 7.33 & - & - & - & - & -\\
		~ & LayoutFormer++ \cite{layouttfpp2023jiang} & CVPR'23 & \textcolor{yellow1}{E-D} & $\usym{2718}$ & $\usym{2718}$ & \textbf{0.72} & 0.34 & \textbf{3.70} & \underline{0.47} & \textbf{1.50} & \textbf{0.76} & 2.89 & \underline{0.62} & 4.00 & - & - & - & - & -\\
		~ & LayoutNUWA-DS \cite{layoutnuwa2024tang} & ICLR'24 & \textcolor{blue1}{Dec} & $\usym{2718}$ & $\usym{2718}$ & 4.02 & 0.12 & 16.10 & 0.47 & 4.50 & 2.98 & 0.10 & 0.47 & 4.67 & \textbf{5.36} & - & - & 0.35 & 2.50\\
		\cline{5-20}
		~ & LayoutDM \cite{layoutdm2023Inoue} & CVPR'23 & \textcolor{red1}{Enc} & $\usym{2713}$ & $\usym{2718}$ & 4.25 & 0.12 & 19.12 & 0.38 & 6.00 & 2.22 & 0.17 & 0.39 & 5.00 & - & - & - & - & -\\
		~ & LDGM \cite{LDGM2023hui} & CVPR'23 & \textcolor{red1}{Enc} & $\usym{2713}$ & $\usym{2718}$ & 19.02 & 0.16 & 21.28 & 0.44 & 7.25 & 12.59 & 0.35 & \underline{0.62} & 6.00 & 17.65 & \textbf{0.45} & 44.25 & 0.37 & 2.50\\
		~ & LayoutPrompter \cite{layoutprompter2023lin} & NeurIPS'23 & \textcolor{blue1}{Dec} & $\usym{2713}$ & $\usym{2718}$  & \underline{1.07} & 0.70 & \underline{9.10} & 0.45 & 4.25 & \underline{1.46} & 2.07 & 0.55 & 5.00 & - & - & - & - & - \\
		~ & LayoutNUWA-DA \cite{layoutnuwa2024tang} & ICLR'24 & \textcolor{blue1}{Dec} & $\usym{2718}$ & $\usym{2713}$ & 3.70 & \textbf{0.03} & 10.80 & \textbf{0.48} & \underline{2.25} & 2.87 & \underline{0.10} & {0.56} & \underline{3.67} & \underline{6.76} & - & - & \textbf{0.42} & \textbf{1.50}\\
		\rowcolor{lightblue}
		~ & LGGPT (Ours) & This Work & \textcolor{blue1}{Dec} & $\usym{2713}$ & $\usym{2713}$ & 5.40 & \underline{0.08} & 9.16 & 0.43 & \cellcolor{colred}4.50 & 1.53 & \textbf{0.09} & \textbf{0.69} & \cellcolor{colred}\textbf{1.67} & {8.99} & \underline{0.51} & \textbf{30.41} & \underline{0.38} & \cellcolor{colred}\underline{2.00}\\
		\hline
		
		\multirow{5}{*}{Relation} & LayoutGAN++ \cite{layoutganpp2021kikuchi} & ACM MM'21 & \textcolor{red1}{Enc} & $\usym{2718}$ & $\usym{2718}$ & 31.87 & 0.21 & 34.39 & 0.38 & 4.00 & 38.89 & 0.54 & 0.38 & 4.33 & 33.88 & 0.59 & 59.43 & 0.27 & \underline{3.00}\\
		~ & LayoutFormer++ \cite{layouttfpp2023jiang} & CVPR'23 & \textcolor{red1}{Enc} & $\usym{2718}$ & $\usym{2718}$ & \underline{4.95} & 0.36 & \underline{7.60} & 0.35 & 3.00 & {5.97} & 4.74 & 0.42 & 3.67 & - & - & - & - & -\\
		\cline{5-20}
		~ & LDGM \cite{LDGM2023hui} & CVPR'23 & \textcolor{red1}{Enc} & $\usym{2713}$ & $\usym{2718}$ & 19.54 & \underline{0.16} & 21.28 & \textbf{0.44} & \underline{2.75} & 16.98 & \underline{0.39} & \textbf{0.61} & \underline{2.33} & \underline{20.58} & \textbf{0.48} & \underline{47.27} & \textbf{0.39} & \textbf{1.50}\\
		~ & LayoutPrompter \cite{layoutprompter2023lin} & NeurIPS'23 & \textcolor{blue1}{Dec} & $\usym{2713}$ & $\usym{2718}$ & \textbf{3.62} & 0.53 & 16.10 & 0.35 & 3.25 & \underline{5.18} & 1.44 & 0.40 & 3.33 & - & - & - & - & -\\
		\rowcolor{lightblue}
		~ & LGGPT (Ours) & This Work & \textcolor{blue1}{Dec} & $\usym{2713}$ & $\usym{2713}$ & 6.49 & \textbf{0.06} & \textbf{6.51} & \underline{0.39} & \cellcolor{colred}\textbf{1.75} & \textbf{2.63} & \textbf{0.13} & \underline{0.51} & \cellcolor{colred}\textbf{1.33} & \textbf{9.54} & \underline{0.49} & \textbf{33.74} & \underline{0.38} & \cellcolor{colred}\textbf{1.50}\\
		\hline
		
		\multirow{7}{*}{Refinement} & RUITE \cite{ruite2021rahman} & IUI'21 & \textcolor{red1}{Enc} & $\usym{2718}$ & $\usym{2718}$ & 6.39 & - & - & 0.42 & 5.50 &  3.23 & - & 0.42 & 6.00 & - & - & - & - & -\\
		~ & LayoutFormer++ \cite{layouttfpp2023jiang} & CVPR'23 & \textcolor{yellow1}{E-D} & $\usym{2718}$ & $\usym{2718}$ & \textbf{0.09} & 0.34 & \textbf{0.60} & \textbf{0.79} & \textbf{1.75} & \textbf{0.03} & 1.76 & \textbf{0.82} & \textbf{2.00} & - & - & - & - & -\\
		~ & LayoutDiffusion \cite{layoutdiff2023zhang} & ICCV'23 & \textcolor{red1}{Enc} & $\usym{2718}$ & $\usym{2718}$ & 2.05 & \textbf{0.04} & \underline{0.79} & \underline{0.66} & \underline{2.00} & 0.55 & \textbf{0.10} & 0.72 & \underline{2.67} & - & - & - & - & -\\
		\cline{5-20}
		~ & LayoutDM \cite{layoutdm2023Inoue} & CVPR'23 & \textcolor{red1}{Enc} & $\usym{2713}$ & $\usym{2718}$ & 6.75 & - & - & 0.35 & 6.50 & 2.77 & - & 0.37 & 6.00 & - & - & - & - & -\\
		~ & LDGM \cite{LDGM2023hui} & CVPR'23 & \textcolor{red1}{Enc} & $\usym{2713}$ & $\usym{2718}$ & 15.28 & 0.10 & 13.05 & 0.48 & 5.00 & 13.19 & {0.33} & 0.62 & 5.00 & 14.95 & {0.42} & {37.22} & 0.39 & {2.00}\\
		~ & LayoutPrompter \cite{layoutprompter2023lin} & NeurIPS'23 & \textcolor{blue1}{Dec} & $\usym{2713}$ & $\usym{2718}$ & \underline{0.28} & 1.03 & 4.80 & 0.65 & 3.50 & 0.98 & 2.27 & 0.75 & 4.00 & - & - & - & - & -\\
		\rowcolor{lightblue}
		~ & LGGPT (Ours) & This Work & \textcolor{blue1}{Dec} & $\usym{2713}$ & $\usym{2713}$ & {0.33} & \underline{0.07} & 4.05 & \underline{0.66} & \cellcolor{colred}{2.50} & \underline{0.52} & \underline{0.14} & \underline{0.77} & \cellcolor{colred}\textbf{2.00} & \textbf{7.86} & \textbf{0.40} & \textbf{28.07} & \textbf{0.48} & \cellcolor{colred}\textbf{1.00}\\
		\hline
		
		\multirow{9}{*}{Gen-U} & LayoutTransformer \cite{layouttf2021gupta} & ICCV'21 & \textcolor{blue1}{Dec} & $\usym{2718}$ & $\usym{2718}$ & {13.90} & {0.13} & 10.10 & - & 4.33 & {7.63} & \textbf{0.07} & - & 3.50 & - & - & - & - & -\\
		~ & BLT \cite{blt2022kong} & ECCV'22 & \textcolor{red1}{Enc} & $\usym{2718}$ & $\usym{2718}$ & 116.00 & 0.15 & 96.00 & - & 7.00 & 88.20 & 1.03 & - & 9.00 & - & - & - & - & -\\
		~ & LayoutFormer++ \cite{layouttfpp2023jiang} & CVPR'23 & \textcolor{yellow1}{E-D} & $\usym{2718}$ & $\usym{2718}$ & 46.52 & 0.41 & \textbf{0.09} & \underline{0.42} & 5.00 & 19.69 & 0.67 & \textbf{0.74} & 6.00 & - & - & - & - & -\\
		~ & LayoutDiffusion \cite{layoutdiff2023zhang} & ICCV'23 & \textcolor{red1}{Enc} & $\usym{2718}$ & $\usym{2718}$ & \underline{8.63} & \underline{0.07} & \underline{0.30} & \underline{0.42} & 1.75 & \textbf{2.49} & \textbf{0.07} & 0.62 & \textbf{1.67} & - & - & - &- & -\\
		~ & LayoutNUWA-DS \cite{layoutnuwa2024tang} & ICLR'24 & \textcolor{blue1}{Dec} & $\usym{2718}$ & $\usym{2718}$ & 8.91 & 0.10 & - & - & \underline{3.00} & 5.67 & 0.12 & - & 3.50 & 24.11 & 1.30 & - & - & 3.00\\
		\cline{5-20}
		~ & LayoutDM \cite{layoutdm2023Inoue} & CVPR'23 & \textcolor{red1}{Enc} & $\usym{2713}$ & $\usym{2718}$ & 13.90 & 0.20 & 18.80 & - & 5.67 & {6.65} & 0.16 & - & 4.50 & - & - & - & - & -\\
		~ & LDGM \cite{LDGM2023hui} & CVPR'23 & \textcolor{red1}{Enc} & $\usym{2713}$ & $\usym{2718}$ & 25.94 & 0.25 & 19.83 & \textbf{0.46} & 5.50 & 26.06 & 0.36 & 0.62 & 6.00 & 32.73 & \textbf{0.47} & 46.43 & \textbf{0.38} & \underline{2.00}\\
		~ & LayoutNUWA-DA \cite{layoutnuwa2024tang} & ICLR'24 & \textcolor{blue1}{Dec} & $\usym{2718}$ & $\usym{2713}$ & 9.21 & 0.18 & - & - & 5.00 & 6.93 & 0.20 & - & 5.50 & \underline{28.93} & 1.03 & - & - & 3.00\\
		\rowcolor{lightblue}
		~ & LGGPT (Ours) & This Work & \textcolor{blue1}{Dec} & $\usym{2713}$ & $\usym{2713}$ & \textbf{7.21} & \textbf{0.06} & 2.74 & \underline{0.42} & \cellcolor{colred}\textbf{1.75} & \underline{2.55} & \underline{0.10} & \underline{0.66} & \cellcolor{colred}\underline{2.33} & \textbf{8.90} & \underline{0.55} & \textbf{27.42} & \textbf{0.38} & \cellcolor{colred}\textbf{1.67}\\
		\hline
		
		\scriptsize \emph{Hybrid tasks}\\
		\hline
		~ & LDGM (Gen-PCM) & CVPR'23 & \textcolor{red1}{Enc} & $\usym{2713}$ & $\usym{2718}$ & 25.76 & \textbf{0.14} & 19.68 & 0.42 & \underline{1.75} & 21.59 & 0.40 & 0.59 & \underline{2.00} & 24.45 & \textbf{0.49} & {44.41} & 0.37 & {1.75}\\
		\rowcolor{lightblue}
		\multirow{-2}{*}{Comp.-Refine.} & LGGPT (Ours) & This Work & \textcolor{blue1}{Dec} & $\usym{2713}$ & $\usym{2713}$ & \textbf{3.88} & 0.17 & \textbf{9.58} & \textbf{0.47} & \cellcolor{colred}\textbf{1.25} & \textbf{2.37} & \textbf{0.18} & \textbf{0.68} & \cellcolor{colred}\textbf{1.00} & \textbf{11.91} & 0.62 & \textbf{36.67} & \textbf{0.43} & \cellcolor{colred}\textbf{1.25}\\
		\hline
		
		~ & LDGM (Gen-CM) & CVPR'23 & \textcolor{red1}{Enc} & $\usym{2713}$ & $\usym{2718}$ & 24.94 & \textbf{0.11} & 16.26 & \textbf{0.44} & \textbf{1.50} & 26.15 & 0.38 & \textbf{0.57} & \underline{1.67} & \textbf{28.74} & \textbf{0.51} & 43.25 & 0.37 & \textbf{1.25}\\
		\rowcolor{lightblue}
		\multirow{-2}{*}{Gen-PS-Refine.} & LGGPT (Ours) & This Work & \textcolor{blue1}{Dec} & $\usym{2713}$ & $\usym{2713}$ & \textbf{13.44} & 0.20 & \textbf{12.15} & 0.37 & \cellcolor{colred}\textbf{1.50} & \textbf{12.47} & \textbf{0.29} & 0.51 & \cellcolor{colred}\textbf{1.33} & {32.48} & 0.89 & \textbf{33.52} & \textbf{0.30} & \cellcolor{colred}1.75\\
		\hline
		
		~ & LDGM (Gen-PM) & CVPR'23 & \textcolor{red1}{Enc} & $\usym{2713}$ & $\usym{2718}$ & 23.58 & 0.10 & 14.11 & 0.46 & \underline{2.00} & 21.64 & 0.38 & 0.58 & \underline{2.00} & 27.33 & \textbf{0.47} & {39.02} & 0.38 & {1.75}\\
		\rowcolor{lightblue}
		\multirow{-2}{*}{Gen-TSP} & LGGPT (Ours) & This Work & \textcolor{blue1}{Dec} & $\usym{2713}$ & $\usym{2713}$ & \textbf{3.28} & \textbf{0.06} & \textbf{7.43} & \textbf{0.49} & \cellcolor{colred}\textbf{1.00} &\textbf{1.20} & \textbf{0.10} & \textbf{0.72} & \cellcolor{colred}\textbf{1.00} & \textbf{8.78} & 0.50 & \textbf{28.63} & \textbf{0.43} & \cellcolor{colred}\textbf{1.25}\\
		\hline
		
		~ & LDGM (Gen-PCM) & CVPR'23 & \textcolor{red1}{Enc} & $\usym{2713}$ & $\usym{2718}$ & 25.76 & \textbf{0.14} & 19.68 & 0.42 & \underline{1.75} & 21.59 & 0.40 & {0.59} & \underline{2.00} & 24.45 & \textbf{0.49} & {44.41} & 0.37 & {1.75}\\
		\rowcolor{lightblue}
		\multirow{-2}{*}{Gen-Arb-Refine.} & LGGPT (Ours) & This Work & \textcolor{blue1}{Dec} & $\usym{2713}$ & $\usym{2713}$ & \textbf{5.83} & 0.19 & \textbf{12.24} & \textbf{0.45} & \cellcolor{colred}\textbf{1.25} & \textbf{3.07} & \textbf{0.19} & \textbf{0.63} & \cellcolor{colred}\textbf{1.00} & \textbf{15.75} & 0.62 & \textbf{38.93} & \textbf{0.42} & \cellcolor{colred}\textbf{1.25}\\
		\noalign{\vspace{-2pt}}
		\bottomrule
	\end{tabular}}
\end{table*}

\subsection{Implementation Detail}
\textbf{Data preprocessing.} We first standardize the layout element labels across all datasets to lowercase (\emph{e.g.}, unify \verb+"+Text\verb+"+ and \verb+"+text\verb+"+ to \verb+"+text\verb+"+) for universal text representation, and then merge all labels and purge the duplicates. We proportionally scale the width and height of distinct types of layouts while maintaining their aspect ratios, with the longer side being constrained to 1024 and the shorter side scaled accordingly. Elements within layouts are also resized in the same proportion. This normalization serves to mitigate any interpretative challenges for the model that might arise from varying layout sizes, and accommodates the use of IQE. Here, the $l_m$ in Eq.~\ref{eq::iqe} is set to 1024.

\textbf{Data sampling and task sampling strategies.} Our model is trained with all tasks on the amalgamated training data from the four datasets simultaneously, \emph{i.e.}, under both the domain-generic and task-generic settings. For data sampling, we sample the four types of data: article, App UI, magazine, and slide at a ratio of 1:7:95:111 (reciprocal of the training data amount ratio) to balance their volume. For task sampling, we categorize the tasks into two broad categories according to task characteristics: Mixed Generation and Single-Type Generation, assigning sampling ratios of 75\% and 25\%, respectively. Detailed specifications for each category are as follows:

\noindent\textbf{Mixed Generation Tasks (75\%):}
\begin{itemize}
	\item \textbf{Mixed Generation without Refinement (45\%).} This involves simulating layout generation from partial layouts by assigning only the \emph{unknown} state to certain attributes, as mentioned in Sec.~\ref{sec::ali}.
	\item \textbf{Mixed Generation with Refinement (30\%).} Both \emph{unknown} and \emph{noisy} states are assigned to attributes, requiring the model to not only generate complete layouts but also refine the noisy elements.
\end{itemize}
Among the Mixed Generation, we randomly incorporate layout relations ([\textbf{Relation}] as detailed in Fig.~\ref{Fig::arch} and Sec.~\ref{sec::ali}) 20\% of the time, simulating the Relationship task. The training on mixed generation tasks primarily contributes to model’s ability to handle arbitrary layout conditions.

\noindent\textbf{Single-Type Generation Tasks (25\%):}
\begin{itemize}
	\item \textbf{Refinement (10\%).} This task demands denoising the entire but not partial layout sequence, where all layout attributes are provided to the model.
	\item \textbf{Gen-U and Gen-UP (7.5\% each).} These tasks involve generating layouts from scratch, treated as specific individual tasks due to their distinct nature.
\end{itemize}
These tasks require specific processing during training thus they are handled separately.

\textbf{Optimization.} We implement LGGPT using Pytorch \cite{pytorch2019paszke}. We utilize the LLM implementations and corresponding tokenizers from Hugging Face Transformers \cite{huggingface2020wolf}. To expedite training, we use a series of accelerating techniques, such as DeepSpeed framework \cite{deepspeed2020rasley}, brain float 16-bit data type. We train the models using 8 NVIDIA A6000 GPUs, with a batch size of 36 and 23,000 training steps for around 2 days. We adopt the AdamW \cite{adamw2019loshchilov} optimizer with $\beta_1=0.9$ and $\beta_2=0.999$ for model training. The learning rate is initially set to 0.0001 and descended according to the cosine schedule. The weight decay rate is set to 0.01. We tokenize inputs using the byte-pair-encoding (BPE) \cite{bpe2016rico}.

\begin{table*}[ht]
	\renewcommand{\arraystretch}{1.05}
	\caption{Comparison of training LGGPT on the domain-generic setting and domain-specific setting. Align. denotes the \emph{Alignment} metric. $\downarrow$ signifies that smaller values are better, whereas $\uparrow$ represents the contrast.}
	\label{Table::comp_separate_train}
	\centering
	\resizebox*{\linewidth}{!}{
		\begin{tabular}{c c c c c c c c c c c c c c}
			\toprule
			\multirow{2.4}{*}{Task} & \multirow{2.4}{*}{Method} & \multirow{2.4}{*}{Domain} &  \multicolumn{4}{c}{PubLayNet \cite{publaynet2019zhong}} & \multicolumn{3}{c}{Rico \cite{rico2017deka}} & \multicolumn{4}{c}{Magazine \cite{magazine2019zheng}}\\
			\cmidrule(r){4-7}\cmidrule(r){8-10}\cmidrule{11-14}
			~ & ~ & ~ & FID $\downarrow$ & Align. $\downarrow$ & Overlap $\downarrow$ & Max IOU $\uparrow$ & FID $\downarrow$ & Align. $\downarrow$ & Max IOU $\uparrow$ & FID $\downarrow$ & Align. $\downarrow$ & Overlap $\downarrow$ & Max IOU $\uparrow$\\
			\hline
			\scriptsize \emph{Singe tasks}\\
			\hline
			
			\multirow{2}{*}{Completion} & LGGPT & Specific & \textbf{1.47} & \textbf{0.03} & \textbf{2.31} & \textbf{0.60} & 1.45 & \textbf{0.09} & 0.78 & 9.63 & \textbf{0.41} & 44.21 & 0.41\\
			~ & LGGPT & Generic & {2.08} & 0.04 & 5.54 & 0.57 & \textbf{1.03} & \textbf{0.12} & \textbf{0.80} & \textbf{8.11} & {0.44} & \textbf{30.60} & \textbf{0.47}\\
			\hline
			
			\multirow{2}{*}{Gen-T} & LGGPT & Specific & \textbf{5.79} & \textbf{0.04} & \textbf{2.49} & \textbf{0.41} & \textbf{2.39} & \textbf{0.09} & \textbf{0.63} & {10.04} & \textbf{0.45} & {40.51} & {0.35}\\
			~ & LGGPT & Generic & {5.94} & 0.06 & 5.20 & \textbf{0.41} & 2.45 & {0.10} & {0.61} & \textbf{9.33} & {0.46} & \textbf{31.58} & \textbf{0.36}\\
			\hline
			
			\multirow{2}{*}{Gen-TS} & LGGPT & Specific & \textbf{4.02} & \textbf{0.06} & \textbf{4.95} & \textbf{0.45} & {1.85} & \textbf{0.07} & \textbf{0.69} & 10.79 & {0.52} & 43.77 & 0.34\\
			~ & LGGPT & Generic & 5.40 & {0.08} & 9.16 & 0.43 & \textbf{1.53} & {0.09} & \textbf{0.69} & \textbf{8.99} & \textbf{0.51} & \textbf{30.41} & \textbf{0.38}\\
			\hline
			
			\multirow{2}{*}{Relation} & LGGPT & Specific & \textbf{5.70} & \textbf{0.05} & \textbf{2.97} & \textbf{0.40} & {2.68} & \textbf{0.06} & \textbf{0.59} & 11.28 & 0.51 & 41.83 & 0.33\\
			~ & LGGPT & Generic & 6.49 & {0.06} & {6.51} & {0.39} & \textbf{2.63} & {0.13} & {0.51} & \textbf{9.54} & \textbf{0.49} & \textbf{33.74} & \textbf{0.38}\\
			\hline
			
			\multirow{2}{*}{Refinement} & LGGPT & Specific & 0.62 & \textbf{0.07} & \textbf{2.41} & {0.65} & 1.06 & \textbf{0.13} & 0.75 & \textbf{7.90} & 0.43 & 44.13 & \textbf{0.48}\\
			~ & LGGPT & Generic & \textbf{0.33} & \textbf{0.07} & 4.05 & \textbf{0.66} & \textbf{0.52} & {0.14} & \textbf{0.77} & {7.86} & \textbf{0.40} & \textbf{28.07} & \textbf{0.48}\\
			\hline
			
			\multirow{2}{*}{Gen-U} & LGGPT & Specific & {7.58} & \textbf{0.05} & \textbf{1.25} & \textbf{0.42} & {3.58} & \textbf{0.07} & \textbf{0.68} & 11.63 & \textbf{0.34} & 40.23 & 0.29\\
			~ & LGGPT & Generic & \textbf{7.21} & {0.06} & 2.74 & \textbf{0.42} & \textbf{2.55} & {0.10} & {0.66} & \textbf{8.90} & {0.55} & \textbf{27.42} & \textbf{0.38}\\
			\hline
			
			\multirow{2}{*}{Gen-UP} & LGGPT & Specific & {7.56} & \textbf{0.05} & \textbf{1.55} & \textbf{0.42} & \textbf{3.70} & 0.04 & 0.68 & 10.67 & \textbf{0.41} & 43.81 & 0.32\\
			~ & LGGPT & Generic & \textbf{7.11} & \textbf{0.06} & 2.83 & \textbf{0.42} & 9.27 & \textbf{0.05} & \textbf{0.84} & \textbf{10.40} & {0.74} & \textbf{31.34} & \textbf{0.36}\\
			\hline
			
			\scriptsize \emph{Hybrid tasks}\\
			\hline
			
			\multirow{2}{*}{Comp.-Refine.} & LGGPT & Specific & {4.15} & 0.20 & \textbf{7.51} & \textbf{0.49} & \textbf{2.30} & \textbf{0.12} & 0.67 & \textbf{11.84} & \textbf{0.62} & 49.23 & 0.42\\
			~ & LGGPT & Generic & \textbf{3.88} & \textbf{0.17} & {9.58} & {0.47} & {2.37} & {0.18} & \textbf{0.68} & {11.91} & \textbf{0.62} & \textbf{36.67} & \textbf{0.43}\\
			\hline
			
			\multirow{2}{*}{Gen-PS-Refine.} & LGGPT & Specific & \textbf{8.15} & {0.21} & \textbf{9.78} & \textbf{0.41} & \textbf{5.36} & \textbf{0.22} & {0.54} & \textbf{16.11} & \textbf{0.51} & 48.49 & \textbf{0.34}\\
			~ & LGGPT & Generic & {13.44} & \textbf{0.20} & {12.15} & 0.37 & {12.47} & {0.29} & \textbf{0.51} & {32.48} & 0.89 & \textbf{33.52} & {0.30}\\
			\hline
			
			\multirow{2}{*}{Gen-TSP} & LGGPT & Specific & \textbf{2.37} & \textbf{0.04} & \textbf{3.55} & \textbf{0.52} & {1.66} & \textbf{0.09} & \textbf{0.72} & 10.19 & \textbf{0.49} & 44.30 & 0.38\\
			~ & LGGPT & Generic & {3.28} & {0.06} & {7.43} & {0.49} & \textbf{1.20} & {0.10} & \textbf{0.72} & \textbf{8.78} & {0.50} & \textbf{28.63} & \textbf{0.43}\\
			\hline
			
			\multirow{2}{*}{Gen-Arb-Refine.} & LGGPT & Specific & \textbf{5.29} & \textbf{0.19} & \textbf{9.21} & \textbf{0.47} & {3.12} & {0.20} & \textbf{0.64} & \textbf{14.47} & \textbf{0.58} & 51.23 & 0.37\\
			~ & LGGPT & Generic & {5.83} & \textbf{0.19} & {12.24} & {0.45} & \textbf{3.07} & \textbf{0.19} & {0.63} & {15.75} & {0.62} & \textbf{38.93} & \textbf{0.42}\\
			\noalign{\vspace{-2pt}}
			\bottomrule
	\end{tabular}}
\end{table*}

\subsection{Comparison with State-of-the-Art Methods}
\label{exp::comp_sota}
We compare our proposed LGGPT to state-of-the-art (SOTA) methods on isolated datasets. Note that our methods are trained simultaneously on all tasks and all domains of data while being tested separately on specific datasets. We use the domain-agnostic version of LayoutNUWA \cite{layoutnuwa2024tang} to mirror our domain-generic setting, which is trained with article, App UI, and magazine data in tandem but tested separately. We denote it as LayoutNUWA-DA. The domain-specific version is also compared, denoted as LayoutNUWA-DS. All other methods are trained and tested in the domain-specific manner. For the Rico \cite{rico2017deka} dataset, the inherent overlap among layout elements makes the Overlap metric unsuitable for gauging the quality of generated layouts. Hence we abstain from computing this metric in the evaluation on Rico.

Furthermore, due to the inconsistent dimensions and different importance of the metrics used, it is infeasible to unify them through linear weighting. Therefore, we specifically design a \emph{Ranking Score} to provide a more intuitive and overall demonstration of the models' performance rankings, which are calculated by averaging the ranking of different metrics. For example, on the PubLayNet dataset, the Ranking Score of a specific model for each task is calculated as $(R_{FID} + R{Align.} + R_{Overlap} + R_{Max IOU}) / 4$, and so forth. The results are summarized in Table~\ref{Table::isolated_comp}.

From assessments on six separate tasks, we have the following observations. (1) In terms of metric performance, LGGPT achieves top-tier results, demonstrating the best Ranking Scores in most cases or maintaining a close second. It is particularly evident in FID and Max IOUs for tasks like Completion, Relation, and Gen-U tasks. Better FID indicates higher generation fidelity, which is conducive to generating layouts that naturally conform to human visual delight. Max IOU is an effective measurement of the adherence of input and output layouts as detailed in Sec.~\ref{exp::metric}. Therefore, the prominent Max IOU of LGGPT demonstrates its better capability in preserving input requirements without being altered, providing enhanced user experience in practice. (2) Compared to prior layout LLM models, \emph{i.e.} LayoutPrompter \cite{layoutprompter2023lin} and LayoutNUWA \cite{layoutnuwa2024tang}, LGGPT surpasses them either in the overall Ranking Scores or the separate metrics. Notably, LGGPT has far fewer parameters (1.5B vs 175B \cite{layoutprompter2023lin} or 7B \cite{layoutnuwa2024tang}) and is pretraind on much more limited pretraining data. The outperformance could be attributed to the tailored ALI and IQE strategy, which convey condensed layout knowledge with a more compact and informative structure. They better facilitate LLM to learn effective task-generic and domain-generic signals through instruction tuning, unleashing its innate reasoning expertise for improved performance. (3) Comparing the most generic LGGPT with other partial-generic or non-generic models, LGGPT is mostly beaten by task-specific models, in which LayoutFormer++ \cite{layouttfpp2023jiang} holds sway in many cases, even yielding some exceptional results (\emph{e.g.}, the Overlap of task Gen-U on PubLayNet and the FID of task Gen-TS on Rico). Nevertheless, it is a task-specific and domain-specific method trained separately for each task and each domain. This singular focus significantly simplifies the training process compared to the more complex task-generic and domain-generic approach of LGGPT, reasonably rendering its outperformances. Despite the more rigorous training regimen, LGGPT still surpasses other state-of-the-art task-generic models like LayoutDM \cite{layoutdm2023Inoue} and LDGM \cite{LDGM2023hui} in most cases, underscoring the superiority of LGGPT as a generic layout generation model.


For comparison on hybrid tasks, we align our optimized definitions with the vanilla settings set forth by LDGM \cite{LDGM2023hui} to perform comparisons. Except for LDGM, none of the other compared methods contemplate evaluating these more complicated tasks. From Table~\ref{Table::isolated_comp}, LGGPT exceeds LDGM by large margins in most cases. It is noteworthy that our LGGPT is trained in a domain-generic manner, which presents greater challenges compared to training on single-domain layout data (detailed in the next subsection). This suggests that LGGPT is impressively more effective in handling arbitrary generation conditions that span multiple domains, underscoring its versatility in meeting variable user requirements in real-world applications. To sum up, despite the substantial challenge derived from the broadest unification setting of LGGPT, it attains competitive or better performance than existing methods on either single-task or hybrid-task comparisons, reinforcing its potency and broad applicability.

\begin{table*}[h]
	\renewcommand{\arraystretch}{1.1}
	\caption{Ablation study on the Interval Quantization Encoding (IQE) strategy. The baseline denotes the scheme of placing placeholder tokens ``$\_$" on the positions of attributes in the \emph{unknown} status. Experiments are conducted on the PubLayNet \cite{publaynet2019zhong} dataset.}
	\label{Table::abl_iqe1}
	\centering
	\resizebox*{0.75\linewidth}{!}{
		\begin{tabular}{c c | c c c c | c c c c}
			\toprule
			\multirow{2.5}{*}{Mode} &\multirow{2.5}{*}{Task/Dataset} & \multicolumn{4}{c|}{Baseline} & \multicolumn{4}{c}{IQE}\\
			\cmidrule(r){3-6}\cmidrule(r){7-10}
			~ & ~ & FID $\downarrow$ & Align. $\downarrow$ & Overlap $\downarrow$ & Max IOU $\uparrow$ & FID $\downarrow$ & Align. $\downarrow$ & Overlap $\downarrow$ & Max IOU $\uparrow$\\
			\midrule
			
			\multirow{7}{*}{Separate} & Completion & 27.87 & 0.13 & 19.86 & 0.27 & \textbf{2.08} & \textbf{0.04} & \textbf{5.54} & \textbf{0.57}\\
			
			~ & Gen-T & 41.89 & 0.23 & 12.95 & 0.45 & \textbf{5.94} & \textbf{0.06} & \textbf{5.20} & \textbf{0.41}\\
			
			~ & Gen-TS & 67.32 & 0.56 & {9.39} & 0.27 & \textbf{5.40} & \textbf{0.08} & \textbf{9.16} & \textbf{0.43}\\
			
			~ & Relation & 10.00 & 0.08 & 21.80 & 0.37 & \textbf{6.49} & \textbf{0.06} & \textbf{6.51} & \textbf{0.39}\\
			
			~ & Refinement & 0.55 & \textbf{0.07} & 8.99 & \textbf{0.66} & \textbf{0.33} & \textbf{0.07} & \textbf{4.05} & \textbf{0.66}\\
			
			~ & Gen-U & 19.12 & 0.08 & 14.22 & 0.40 & \textbf{7.21} & \textbf{0.06} & \textbf{2.74} & \textbf{0.42}\\
			
			~ & Gen-UP & 7.49 & 0.07 & 15.00 & \textbf{0.42} & \textbf{7.11} & \textbf{0.06} & \textbf{2.83} & 0.41\\
			\midrule
			
			\multirow{4}{*}{Hybrid} & Comp.-Refine. & 18.40 & \textbf{0.16} & 16.79 & 0.40 & \textbf{3.88} & 0.17 & \textbf{9.58} & \textbf{0.47}\\
			
			~ & Gen-PS-Refine. & 61.87 & 0.75 & \textbf{15.76} & 0.33 & \textbf{13.44} & \textbf{0.20} & \textbf{12.15} & \textbf{0.37}\\
			
			~ & Gen-TSP & 24.65 & 0.14 & 26.25 & 0.30 & \textbf{3.28} & \textbf{0.06} & \textbf{7.43} & \textbf{0.49}\\
			
			~ & Gen-Arb-Refine. & 13.20 & \textbf{0.19} & 22.05 & 0.41 & \textbf{5.83} & \textbf{0.19} & \textbf{12.24} & \textbf{0.45}\\
			\cmidrule(r){1-6}\cmidrule(r){7-10}
			
			\multirow{2}{*}{\#Avg. Token} & PubLayNet & \multicolumn{4}{c|}{76} & \multicolumn{4}{c}{54}\\
			~ & Rico & \multicolumn{4}{c|}{114} & \multicolumn{4}{c}{79}\\
			\bottomrule
	\end{tabular}}
\end{table*}

\begin{table*}[h]
	\renewcommand{\arraystretch}{1.1}
	\caption{Ablation study on the effectiveness of ALI. We borrow the HTML-based layout format (golden code) from LayoutNUWA \cite{layoutnuwa2024tang} as the input template and compare it against our proposed ALI. The experiments are conducted on the PubLayNet \cite{publaynet2019zhong} dataset.}
	\label{Table::abl_html}
	\centering
	\resizebox*{0.75\linewidth}{!}{
		\begin{tabular}{c c | c c c c | c c c c}
			\toprule
			\multirow{2.5}{*}{Mode} &\multirow{2.5}{*}{Task} & \multicolumn{4}{c}{HTML-based} & \multicolumn{4}{c}{ALI}\\
			\cmidrule(r){3-6}\cmidrule(r){7-10}
			~ & ~ & FID $\downarrow$ & Align. $\downarrow$ & Overlap $\downarrow$ & Max IOU $\uparrow$ & FID $\downarrow$ & Align. $\downarrow$ & Overlap $\downarrow$ & Max IOU $\uparrow$\\
			\midrule
			
			\multirow{7}{*}{Separate} & Completion & 4.12 & 0.05 & 13.21 & 0.56 & \textbf{2.08} & \textbf{0.04} & \textbf{5.54} & \textbf{0.57}\\
			
			~ & Gen-T & 7.32 & 0.08 & 16.63 & 0.40 & \textbf{5.94} & \textbf{0.06} & \textbf{5.20} & \textbf{0.41}\\
			
			~ & Gen-TS & 11.53 & 0.12 & 12.42 & 0.42 & \textbf{5.40} & \textbf{0.08} & \textbf{9.16} & \textbf{0.43}\\
			
			~ & Relation & 8.93 & 0.10 & 18.64 & 0.37 & \textbf{6.49} & \textbf{0.06} & \textbf{6.51} & \textbf{0.39}\\
			
			~ & Refinement & 0.47 & \textbf{0.06} & 5.09 & \textbf{0.68} & \textbf{0.33} & {0.07} & \textbf{4.05} & {0.66}\\
			
			~ & Gen-U & 9.76 & 0.09 & 7.30 & 0.40 & \textbf{7.21} & \textbf{0.06} & \textbf{2.74} & \textbf{0.42}\\
			
			~ & Gen-UP & 10.89 & 0.08 & 6.95 & 0.41 & \textbf{7.11} & \textbf{0.06} & \textbf{2.83} & \textbf{0.41}\\
			\midrule
			
			\multirow{4}{*}{Hybrid} & Comp.-Refine. & 7.49 & \textbf{0.17} & 17.93 & 0.46 & \textbf{3.88} & \textbf{0.17} & \textbf{9.58} & \textbf{0.47}\\
			
			~ & Gen-PS-Refine. & 26.91 & 0.27 & 16.61 & 0.35 & \textbf{13.44} & \textbf{0.20} & \textbf{12.15} & \textbf{0.37}\\
			
			~ & Gen-TSP & 4.51 & 0.07 & 10.71 & \textbf{0.49} & \textbf{3.28} & \textbf{0.06} & \textbf{7.43} & \textbf{0.49}\\
			
			~ & Gen-Arb-Refine. & 8.15 & 0.23 & 14.87 & 0.44 & \textbf{5.83} & \textbf{0.19} & \textbf{12.24} & \textbf{0.45}\\
			
			\midrule
			\multicolumn{2}{c|}{Inference Time Cost/per sample} & \multicolumn{4}{c|}{3.08s} & \multicolumn{4}{c}{1.83s}\\
			
			\bottomrule
	\end{tabular}}
\end{table*}

\begin{table*}[h]
	\renewcommand{\arraystretch}{1.1}
	\caption{Ablation study on the utilization of pretrained weights of the LLM. We train LGGPT from scratch and with the pretrained weights (default) on the PubLayNet \cite{publaynet2019zhong} dataset for comparison.}
	\label{Table::abl_pretrain}
	\centering
	\resizebox*{0.75\linewidth}{!}{
		\begin{tabular}{c c | c c c c | c c c c}
			\toprule
			\multirow{2.4}{*}{Mode} &\multirow{2.4}{*}{Task} & \multicolumn{4}{c}{Without Pretrained Weights} & \multicolumn{4}{c}{With Pretrained Weights}\\
			\cmidrule(r){3-6}\cmidrule(r){7-10}
			~ & ~ & FID $\downarrow$ & Align. $\downarrow$ & Overlap $\downarrow$ & Max IOU $\uparrow$ & FID $\downarrow$ & Align. $\downarrow$ & Overlap $\downarrow$ & Max IOU $\uparrow$\\
			\midrule
			
			\multirow{7}{*}{Separate} & Completion & 3.08 & 0.05 & 9.30 & \textbf{0.58} & \textbf{2.08} & \textbf{0.04} & \textbf{5.54} & {0.57}\\
			
			~ & Gen-T & 6.51 & \textbf{0.06} & 11.54 & 0.40 & \textbf{5.94} & \textbf{0.06} & \textbf{5.20} & \textbf{0.41}\\
			
			~ & Gen-TS & 5.61 & 0.09 & 12.83 & \textbf{0.43} & \textbf{5.40} & \textbf{0.08} & \textbf{9.16} & \textbf{0.43}\\
			
			~ & Relation & \textbf{5.92} & {0.07} & {11.55} & \textbf{0.39} & {6.49} & \textbf{0.06} & \textbf{6.51} & \textbf{0.39}\\
			
			~ & Refinement & 6.75 & {0.09} & 16.89 & {0.53} & \textbf{0.33} & \textbf{0.07} & \textbf{4.05} & \textbf{0.66}\\
			
			~ & Gen-U & 9.16 & 0.07 & {11.82} & {0.40} & \textbf{7.21} & \textbf{0.06} & \textbf{2.74} & \textbf{0.42}\\
			
			~ & Gen-UP & 8.25 & \textbf{0.06} & 8.62 & \textbf{0.41} & \textbf{7.11} & \textbf{0.06} & \textbf{2.83} & \textbf{0.41}\\
			\midrule
			
			\multirow{4}{*}{Hybrid} & Comp.-Refine. & {13.66} & \textbf{0.17} & 17.44 & {0.43} & \textbf{3.88} & \textbf{0.17} & \textbf{9.58} & \textbf{0.47}\\
			
			~ & Gen-PS-Refine. & {19.93} & {0.21} & {16.37} & {0.36} & \textbf{13.44} & \textbf{0.20} & \textbf{12.15} & \textbf{0.37}\\
			
			~ & Gen-TSP & 4.51 & 0.07 & 12.01 & {0.47} & \textbf{3.28} & \textbf{0.06} & \textbf{7.43} & \textbf{0.49}\\
			
			~ & Gen-Arb-Refine. & 14.88 & \textbf{0.18} & {18.59} & 0.41 & \textbf{5.83} & {0.19} & \textbf{12.24} & \textbf{0.45}\\
			\bottomrule
	\end{tabular}}
\end{table*}

\subsection{Comparison on Domain Setting}
\label{exp::domain}
We train LGGPT under the domain-specific setting to gauge its performance against the default domain-generic setting. The results are presented in Table~\ref{Table::comp_separate_train}. For evaluations on single tasks, we observe that in the domain-specific setting, LGGPT yields matched or superior performance compared to the domain-generic approach, particularly evident in the improvement of the Overlap metric when tested on the PubLayNet and Rico datasets. For evaluations on hybrid tasks, the performance of domain-specific LGGPT outstrips its domain-generic counterpart, especially in terms of FID and Max IOU. This performance distinction between the two settings substantiates the greater challenge of domain-generic scenarios, where the huge variation of layout styles across different domains significantly complicates the model's ability to learn the interdependencies of diverse layouts within a unified setting. Despite a larger data volume in the generic setting, it did not translate to performance gains, but rather a decline. This observation aligns with findings from LayoutNUWA \cite{layoutnuwa2024tang}, where domain-specific settings sometimes exceeded the generic ones, such as in the Completion and Gen-U tasks. Nevertheless, even in this more challenging context, LGGPT demonstrates competitive or superior performance when benchmarked against existing methods as demonstrated in Sec.~\ref{exp::comp_sota}, highlighting its robustness and versatility. 

\begin{table*}[t]
	\renewcommand{\arraystretch}{1.1}
	\caption{Comparison between small language models and a large language model (LLM) for unified layout generation. We use LayoutFormer++ \cite{layouttfpp2023jiang} and GPT2-Small as baselines. To ensure a fair comparison, we equip LayoutTransformer++ with our proposed IQE. The experiments are conducted on the PubLayNet \cite{publaynet2019zhong} dataset.}
	\label{Table::abl_transformer1}
	\centering
	\resizebox*{\linewidth}{!}{
		\begin{tabular}{c c | c c c c | c c c c | c c c c}
			\toprule
			\multirow{2.5}{*}{Mode} &\multirow{2.5}{*}{Task} & \multicolumn{4}{c}{LayoutFormer++ w/ IQE (60M)} & \multicolumn{4}{c}{GPT2-Small (137M)} & \multicolumn{4}{c}{GPT2-XL (1.5B)}\\
			\cmidrule(r){3-6}\cmidrule(r){7-10}\cmidrule(r){11-14}
			~ & ~ & FID $\downarrow$ & Align. $\downarrow$ & Overlap $\downarrow$ & Max IOU $\uparrow$ & FID $\downarrow$ & Align. $\downarrow$ & Overlap $\downarrow$ & Max IOU $\uparrow$ & FID $\downarrow$ & Align. $\downarrow$ & Overlap $\downarrow$ & Max IOU $\uparrow$\\
			\midrule
			
			\multirow{7}{*}{Separate} & Completion & 15.23 & 0.09 & 28.74 & 0.39 & 3.33 & 0.05 & 12.21 & 0.55 & \textbf{2.08} & \textbf{0.04} & \textbf{5.54} & \textbf{0.57}\\
			
			~ & Gen-T & 27.03 & 0.15 & 40.60 & 0.33 & 7.33 & 0.07 & 13.43 & 0.40 & \textbf{5.94} & \textbf{0.06} & \textbf{5.20} & \textbf{0.41}\\
			
			~ & Gen-TS & 40.25 & 0.15 & 40.36 & 0.32 & 9.89 & 0.10 & 17.19 & 0.40 & \textbf{5.40} & \textbf{0.08} & \textbf{9.16} & \textbf{0.43}\\
			
			~ & Relation & 25.70 & 0.14 & 45.03 & 0.33 & 8.01 & {0.07} & {14.48} & {0.39} & \textbf{6.49} & \textbf{0.06} & \textbf{6.51} & \textbf{0.39}\\
			
			~ & Refinement & 8.22 & 0.08 & 9.56 & 0.39 & 0.65 & {0.08} & 8.62 & \textbf{0.67} & \textbf{0.33} & \textbf{0.07} & \textbf{4.05} & {0.66}\\
			
			~ & Gen-U & 30.09 & 0.19 & 35.29 & 0.35 & 8.21 & 0.07 & {11.68} & {0.41} & \textbf{7.21} & \textbf{0.06} & \textbf{2.74} & \textbf{0.42}\\
			
			~ & Gen-UP & 28.29 & 0.15 & 36.24 & 0.35 & 8.33 & 0.07 & 11.04 & 0.41 & \textbf{7.11} & \textbf{0.06} & \textbf{2.83} & \textbf{0.41}\\
			\midrule
			
			\multirow{4}{*}{Hybrid} & Comp.-Refine. & 22.62 & 0.24 & 30.81 & 0.35 & {5.53} & {0.18} & 15.12 & {0.46} & \textbf{3.88} & \textbf{0.17} & \textbf{9.58} & \textbf{0.47}\\
			
			~ & Gen-PS-Refine. & 47.78 & 0.30 & 33.61 & 0.27 & {15.71} & \textbf{0.20} & {16.27} & \textbf{0.37} & \textbf{13.44} & \textbf{0.20} & \textbf{12.15} & \textbf{0.37}\\
			
			~ & Gen-TSP & 22.07 & 0.12 & 38.74 & 0.36 & 4.48 & 0.08 & 15.56 & {0.46} & \textbf{3.28} & \textbf{0.06} & \textbf{7.43} & \textbf{0.49}\\
			
			~ & Gen-Arb-Refine. & 29.21 & 0.30 & 39.28 & 0.34 & 8.06 & 0.20 & {17.40} & 0.43 & \textbf{5.83} & \textbf{0.19} & \textbf{12.24} & \textbf{0.45}\\
			\bottomrule
	\end{tabular}}
\end{table*}

Furthermore, this observed performance discrepancy underscores the need for additional research to bridge this gap and improve domain-generic performance. Potential solutions include incorporating data from similar domains for domain-generic training, such as scientific articles and financial documents, which may cultivate a synergistic effect and improve the performance over domain-specific training. Additionally, implementing adaptive loss weighting to specifically optimize different domains in domain-generic training may help mitigate this gap. Exploring more strategies in this direction could be promising for advancing layout generation to achieve both specificity and generality.

\subsection{Ablation Study}
\label{exp::abl}
\subsubsection{Interval Quantization Encoding (IQE) strategy}
We ablate the efficacy of our proposed IQE strategy as shown in Table~\ref{Table::abl_iqe1}. We adopt the most conventional placeholder scheme as the baseline, in which we place the placeholder ``$\_$" on the positions of \emph{unknown} attributes in the input instruction for comparison. First, IQE consistently surpasses the baseline in almost all cases. This superiority is likely due to several aspects. On one hand, IQE preserves valid layout attributes and excludes meaningless symbols such as placeholders. This substantially enriches the information density of layout instructions, thereby facilitating LGGPT to capture essential layout features more effectively. In contrast, the placeholder approach suffers from the fluctuated and variable positions that come with the infinite attribute combinations in the prompt. The model grapples with the unpredictable variation and sparse layout clues to effectively grab layout context, thus rendering considerable performance degradation. Notably, the baseline performs better in tasks where fewer placeholder tokens are used (Refinement, Gen-UP, \emph{etc.}). This phenomenon confirms the inadaptability of using placeholders in scenarios involving highly versatile input sequences, such as unified layout generation.

Second, IQE remarkably reduces the average token number of ALI by about 30\% compared to the baseline. This could expedite LLM's inference in specific cases, such as deploying LLMs on edge devices or other resource-limited scenarios. With the integration of IQE, ALI boasts a more compact structure, which enables LGGPT much better training performance and inference efficiency than former HTML-based layout LLM \cite{layoutprompter2023lin,layoutnuwa2024tang}, further highlighting its qualitative and quantitative advantages.

\subsubsection{Arbitrary Layout Instruction (ALI)}
We compare our proposed ALI against other input templates in the unified layout generation to evaluate its effectiveness, as shown in Table~\ref{Table::abl_html}. We substitute the numerical layout format of our ALI with the HTML layout format (golden code) from LayoutNUWA \cite{layoutnuwa2024tang} to build the layout input, acting as the baseline. From Table~\ref{Table::abl_html}, it can be observed that our ALI outperforms the HTML-based layout template by obvious margins. The HTML code format introduces considerable redundant code symbols to the layout sequence, such as structural descriptors like \texttt{<body>} and \texttt{</body>}, and repeated element descriptions like \texttt{data-category} and \texttt{width}. This redundancy essentially hampers the effective capture of valid layout information, especially in the unified layout generation where layout conditions are highly varying, directly diminishing model performance. In contrast, our ALI features a succinct layout structure that refrains from any redundant descriptions, conveying more condensed layout information to the model and facilitating layout comprehension. Besides, we compare the inference time cost of the HTML-based input template and our ALI in the bottom line of Table~\ref{Table::abl_html}. The excessive code symbols lead to a significant token increase, decelerating the inference from 1.83s to 3.08s per sample. These findings strongly validate the superiority of our proposed ALI as a uniform input template for unified layout generation, both in terms of boosting model performance and accelerating inference.

\subsubsection{The Utilization of Pretrained Weights}
\label{sec::pretrain_weights}
We investigate the efficacy of pretrained weights leveraged in the LLM by training LGGPT with/without the pretrained weights of GPT2-XL. Results are listed in Table~\ref{Table::abl_pretrain}. When trained from scratch, the performance of LGGPT significantly declines compared to training based on the pretrained weights. Apart from the Alignment metric, which is still comparable, other metrics like FID and Overlap grapple with severe degradation. Declines on the more complicated hybrid tasks become much worse than on separate tasks. One of our intentions is to harness the reasoning skills of LLMs, which mainly derive from large-scale pretraining, to address the challenge of task-domain unified layout generation. This outcome justifies that we essentially unleash the reasoning skills of LLM and bring notable performance improvement. Although the layout data format is heterogeneous to LLM (Sec.~\ref{sec::rw_llm}), we demonstrate that these pre-learned reasoning skills can be effectively harnessed through instruction tuning and facilitate the comprehension of complicated layout generation requirements.

\subsection{Necessity of Using an LLM for Unified Layout Generation}
A natural question arises: is a Large Language Model (LLM) truly essential for unified layout generation across both tasks and domains? To explore this, we compare LGGPT (GPT2-XL, 1.5B) with two small language models, \emph{i.e.}, LayoutFormer++ (an established layout generation method based on T5-Small \cite{t52020jmlr}, 60M) and GPT2-Small (137M). For fair comparisons, we apply our proposed IQE technique on LayoutFormer++. The GPT2-Small shares an identical architecture as GPT2-XL, differing only in size due to fewer Transformer layers and reduced embedding size. The results are summarized in Table~\ref{Table::abl_transformer1}. As observed, LayoutTransformer++ delivers much worse performance in the unified scenario. Although it has achieved SOTA results as shown in Table~\ref{Table::isolated_comp}, those pronounced performances come from the separate training on each task and each domain. In the unified training scenario, it struggles with the increased complexity. Also, the smaller GPT2-Small leads to declines in most outcomes. While GPT2-Small occasionally surpasses its larger sibling, the outperformances are relatively marginal.

This experiment substantiates the necessity for employing LLM to solve the unified layout generation. While small language models undoubtedly offer higher training and inference efficiency, this comes at the cost of compromised performance. This shortfall could be ascribed to the complexity of unified learning across both generation tasks and data domains, surpassing the capacity of a small Transformer. Furthermore, since we sought a sweet spot that harmonizes proficiency with resource economy, the GPT2-XL, with its 1.5B parameters, has emerged as a stand-out performer. This is confirmed by its outperformance over the significantly larger 7B or 175B LLMs as detailed in Sec.~\ref{exp::comp_sota} and the comparison with other LLMs presented in Sec.~\ref{sec::effect_llm}.

\begin{table*}[t]
	\renewcommand{\arraystretch}{1.1}
	\caption{Comparison between equal-sized layout-specific model and general LLM for unified layout generation. We compare our LGGPT using GPT2-XL as backbone against LayoutFormer++ \cite{layouttfpp2023jiang} with a 1.5B backbone and its custom tokenizer. For a fair comparison, we also equip this model with our proposed IQE. The experiments are conducted on the PubLayNet \cite{publaynet2019zhong} dataset.}
	\label{Table::comp_similarsize}
	\centering
	\resizebox*{0.75\linewidth}{!}{
		\begin{tabular}{c c | c c c c | c c c c}
			\toprule
			\multirow{2.5}{*}{Mode} &\multirow{2.5}{*}{Task} & \multicolumn{4}{c}{LayoutFormer++ w/ IQE (1.5B)} & \multicolumn{4}{c}{GPT2-XL (1.5B)}\\
			\cmidrule(r){3-6}\cmidrule(r){7-10}
			~ & ~ & FID $\downarrow$ & Align. $\downarrow$ & Overlap $\downarrow$ & Max IOU $\uparrow$ & FID $\downarrow$ & Align. $\downarrow$ & Overlap $\downarrow$ & Max IOU $\uparrow$\\
			\midrule
			
			\multirow{7}{*}{Separate} & Completion & 12.79 & 0.10 & 18.43 & 0.38 & \textbf{2.08} & \textbf{0.04} & \textbf{5.54} & \textbf{0.57}\\
			
			~ & Gen-T & 17.94 & 0.13 & 17.94 & 0.37 & \textbf{5.94} & \textbf{0.06} & \textbf{5.20} & \textbf{0.41}\\
			
			~ & Gen-TS & 16.17 & 0.14 & 18.79 & 0.37 & \textbf{5.40} & \textbf{0.08} & \textbf{9.16} & \textbf{0.43}\\
			
			~ & Relation & 17.63 & 0.14 & 19.27 & 0.36 & \textbf{6.49} & \textbf{0.06} & \textbf{6.51} & \textbf{0.39}\\
			
			~ & Refinement & 14.67 & 0.04 & 22.90 & 0.40 & \textbf{0.33} & \textbf{0.07} & \textbf{4.05} & {0.66}\\
			
			~ & Gen-U & 20.66 & 0.12 & 18.54 & 0.38 & \textbf{7.21} & \textbf{0.06} & \textbf{2.74} & \textbf{0.42}\\
			
			~ & Gen-UP & 17.35 & 0.08 & 17.24 & 0.37 & \textbf{7.11} & \textbf{0.06} & \textbf{2.83} & \textbf{0.41}\\
			\midrule
			
			\multirow{4}{*}{Hybrid} & Comp.-Refine. & 16.43 & 0.20 & 19.01 & 0.37 & \textbf{3.88} & \textbf{0.17} & \textbf{9.58} & \textbf{0.47}\\
			
			~ & Gen-PS-Refine. & 22.27 & 0.33 & 17.27 & 0.35 & \textbf{13.44} & \textbf{0.20} & \textbf{12.15} & \textbf{0.37}\\
			
			~ & Gen-TSP & 13.90 & 0.11 & 17.47 & 0.39 & \textbf{3.28} & \textbf{0.06} & \textbf{7.43} & \textbf{0.49}\\
			
			~ & Gen-Arb-Refine. & 17.32 & 0.25 & 18.81 & 0.37 & \textbf{5.83} & \textbf{0.19} & \textbf{12.24} & \textbf{0.45}\\
			\bottomrule
	\end{tabular}}
\end{table*}

\begin{table*}[t]
	\renewcommand{\arraystretch}{1.1}
	\caption{Comparison of using different scales of LLMs as the core implementation of LGGPT, including GPT2-XL \cite{gpt22019radford} (default), TinyLLaMAv1.1-1.1B \cite{tinyllama2024zhang}, Qwen2-1.5B \cite{qwen22024yang}, Qwen1.5-1.8B \cite{qwen1.52024bai}, NSFW-3B \cite{nsfw}, and LLaMA3-8B \cite{llama32024dubey}. The experiments are conducted on the PubLayNet \cite{publaynet2019zhong} dataset.}
	\label{Table::comp_llm}
	\begin{minipage}{\textwidth}
		\centering
		\resizebox*{\linewidth}{!}{
			\begin{tabular}{c c | c c c c | c c c c | c c c c}
				\toprule
				\multirow{2.5}{*}{Mode} &\multirow{2.5}{*}{Task} & \multicolumn{4}{c}{GPT2-XL} & \multicolumn{4}{c}{Qwen2-1.5B} & \multicolumn{4}{c}{Qwen1.5-1.8B}\\
				\cmidrule(r){3-6}\cmidrule(r){7-10}\cmidrule(r){11-14}
				~ & ~ & FID $\downarrow$ & Align. $\downarrow$ & Overlap $\downarrow$ & Max IOU $\uparrow$ & FID $\downarrow$ & Align. $\downarrow$ & Overlap $\downarrow$ & Max IOU $\uparrow$ & FID $\downarrow$ & Align. $\downarrow$ & Overlap $\downarrow$ & Max IOU $\uparrow$\\
				\midrule
				
				\multirow{7}{*}{Separate} & Completion & \textbf{2.08} & \textbf{0.04} & {5.54} & {0.57} & 2.45 & \textbf{0.04} & 4.63 & 0.57 & 2.88 & 0.05 & 6.74 & 0.59\\
				
				~ & Gen-T & \textbf{5.94} & \textbf{0.06} & {5.20} & \textbf{0.41} & 6.61 & \textbf{0.06} & 4.58 & 0.40 & 7.86 & 0.08 & 8.04 & 0.39\\
				
				~ & Gen-TS & \textbf{5.40} & \textbf{0.08} & {9.16} & {0.43} & 6.55 & 0.09 & \textbf{7.57} & 0.44 & 7.83 & 0.10 & 9.56 & 0.43\\
				
				~ & Relation & \textbf{6.49} & \textbf{0.06} & {6.51} & \textbf{0.39} & 7.18 & 0.07 & 5.37 & 0.39 & 8.80 & 0.08 & 9.73 & 0.38\\
				
				~ & Refinement & \textbf{0.33} & {0.07} & {4.05} & {0.66} & 0.58 & \textbf{0.05} & 1.89 & 0.69 & 0.58 & 0.06 & 2.10 & 0.67\\
				
				~ & Gen-U & \textbf{7.21} & \textbf{0.06} & {2.74} & {0.42} & 7.86 & 0.07 & \textbf{2.47} & 0.41 & 10.28 & 0.10 & 5.18 & 0.41\\
				
				~ & Gen-UP & \textbf{7.11} & \textbf{0.06} & \textbf{2.83} & {0.41} & 7.13 & 0.07 & 3.15 & 0.41 & 8.49 & 0.09 & 6.01 & 0.41\\
				\midrule
				
				\multirow{4}{*}{Hybrid} & Comp.-Refine. & \textbf{3.88} & {0.17} & {9.58} & {0.47} & 4.85 & \textbf{0.16} & \textbf{9.05} & 0.47 & 6.01 & 0.19 & 12.15 & 0.46\\
				
				~ & Gen-PS-Refine. & \textbf{13.44} & \textbf{0.20} & \textbf{12.15} & \textbf{0.37} & 16.63 & 0.23 & 14.13 & 0.36 & 18.70 & 0.25 & 15.71 & 0.36\\
				
				~ & Gen-TSP & \textbf{3.28} & \textbf{0.06} & {7.43} & {0.49} & 3.49 & \textbf{0.06} & 6.42 & 0.50 & 4.39 & 0.07 & 9.22 & 0.49\\
				
				~ & Gen-Arb-Refine. & \textbf{5.83} & {0.19} & \textbf{12.24} & {0.45} & 7.28 & 0.22 & 12.64 & 0.45 & 9.46 & 0.23 & 15.09 & 0.44\\
				\midrule
				\multicolumn{2}{c|}{\#Parameters} & \multicolumn{4}{c|}{1.5B} & \multicolumn{4}{c|}{1.5B} & \multicolumn{4}{c}{1.8B}\\
				\multicolumn{2}{c|}{Inference Time Cost/per sample} & \multicolumn{4}{c|}{\textbf{1.83s}} & \multicolumn{4}{c|}{4.32s} & \multicolumn{4}{c}{3.57s}\\
				\bottomrule
		\end{tabular}}
	\end{minipage}
	\vfill
	\vspace{3pt}
	\begin{minipage}{\textwidth}
		\centering
		\resizebox*{\linewidth}{!}{
			\begin{tabular}{c c | c c c c | c c c c | c c c c}
				\toprule
				\multirow{2.5}{*}{Mode} &\multirow{2.5}{*}{Task} & \multicolumn{4}{c}{TinyLLaMAv1.1-1.1B} & \multicolumn{4}{c}{NSFW-3B} & \multicolumn{4}{c}{LLaMA3-8B}\\
				\cmidrule(r){3-6}\cmidrule(r){7-10}\cmidrule(r){11-14}
				~ & ~ & FID $\downarrow$ & Align. $\downarrow$ & Overlap $\downarrow$ & Max IOU $\uparrow$ & FID $\downarrow$ & Align. $\downarrow$ & Overlap $\downarrow$ & Max IOU $\uparrow$ & FID $\downarrow$ & Align. $\downarrow$ & Overlap $\downarrow$ & Max IOU $\uparrow$\\
				\midrule
				
				\multirow{7}{*}{Separate} & Completion & 2.26 & 0.04 & \textbf{4.09} & \textbf{0.59} & 29.04 & 0.07 & 30.82 & 0.33 & 17.12 & 0.05 & 14.79 & 0.39\\
				
				~ & Gen-T & 6.32 & \textbf{0.06} & \textbf{4.54} & \textbf{0.41} & 35.72 & 0.09 & 31.73 & 0.28 & 52.12 & 0.05 & 11.00 & 0.22\\
				
				~ & Gen-TS & 7.74 & 0.09 & {7.76} & \textbf{0.45} & 32.97 & 0.09 & 32.63 & 0.29 & 60.57 & 0.06 & 15.93 & 0.26\\
				
				~ & Relation & 7.78 & \textbf{0.06} & \textbf{5.36} & \textbf{0.39} & 44.31 & 0.09 & 39.00 & 0.26 & 52.21 & 0.05 & 12.41 & 0.21\\
				
				~ & Refinement & 0.97 & \textbf{0.05} & \textbf{1.57} & \textbf{0.69} & 27.67 & 0.05 & 147.77 & 0.38 & 5.89 & 0.24 & 12.12 & 0.55\\
				
				~ & Gen-U & 8.27 & 0.07 & {2.49} & \textbf{0.43} & 79.92 & 0.07 & 49.23 & 0.23 & 78.63 & 0.02 & 14.60 & 0.24\\
				
				~ & Gen-UP & 7.33 & 0.07 & 3.00 & \textbf{0.42} & 58.17 & 0.07 & 42.20 & 0.27 & 63.74 & 0.04 & 15.68 & 0.24\\
				\midrule
				
				\multirow{4}{*}{Hybrid} & Comp.-Refine. & 5.01 & 0.17 & {9.15} & \textbf{0.48} & 28.92 & 0.14 & 29.38 & 0.32 & 21.94 & 0.18 & 17.92 & 0.36\\
				
				~ & Gen-PS-Refine. & 16.65 & 0.23 & 13.98 & \textbf{0.37} & 44.99 & 0.15 & 33.17 & 0.24 & 50.62 & 0.14 & 18.30 & 0.24\\
				
				~ & Gen-TSP & 3.88 & \textbf{0.06} & \textbf{6.19} & \textbf{0.51} & 31.69 & 0.08 & 32.28 & 0.31 & 33.03 & 0.07 & 17.06 & 0.30\\
				
				~ & Gen-Arb-Refine. & 7.72 & 0.22 & 12.30 & \textbf{0.46} & 33.25 & \textbf{0.14} & 30.80 & 0.31 & 34.12 & \textbf{0.14} & 18.15 & 0.29\\
				\midrule
				\multicolumn{2}{c|}{\#Parameters} & \multicolumn{4}{c|}{1.1B} & \multicolumn{4}{c|}{3B} & \multicolumn{4}{c}{8B}\\
				\multicolumn{2}{c|}{Inference Time Cost/per sample} & \multicolumn{4}{c|}{3.08s} & \multicolumn{4}{c|}{5.76s} & \multicolumn{4}{c}{8.25s}\\
				\bottomrule
		\end{tabular}}
	\end{minipage}
\end{table*}

\subsection{Necessity of Using a General LLM}
Since many transformer-based layout generation models have been proposed, it is worth exploring whether a specialized layout-specific model of the same size could match the performance of a general LLM. Therefore, we scale up LayoutFormer++ \cite{layouttfpp2023jiang} to 1.5B parameters with its custom tokenizer and compared it to our LGGPT based on GPT2-XL. The results are presented in Table 7. As observed, despite having the same parameter size, the layout-specific model significantly underperforms LGGPT. This performance gap highlights the advantages of leveraging general LLMs for unified layout generation. The underperformance of LayoutFormer++ could be ascribed to two key factors: (1) \textbf{Model architecture suitability.} LayoutFormer++ adopts an encoder-decoder architecture, whereas LGGPT uses a decoder-only architecture. Even comparing the performance of both models trained from scratch (Table~\ref{Table::abl_pretrain}), LayoutFormer++ still lags behind LGGPT. This suggests that the encoder-decoder architecture may not be well-suited for unified layout generation, limiting its effectiveness and accounting for its underperformance. (2) \textbf{Pretraining benefits.} General LLMs such as GPT2 benefit from pretrained weights that work seamlessly with their inherent tokenizers. As validated in Sec.~\ref{sec::pretrain_weights}, GPT2-XL with pretrained weights can substantially improve model performance. In contrast, LayoutFormer++ employs a custom tokenizer, preventing it from leveraging the pretrained weights and likely exacerbating its performance deficit. These findings imply that simply scaling up a layout-specific model to a comparable size of general LLMs is insufficient. Leveraging general LLMs for unified layout generation, with the generalized capabilities inherent in their pretrained weights, is crucial for effective unified layout generation.

\subsection{Effect of Using Different LLMs}
\label{sec::effect_llm}
To investigate whether the 1.5B parameters is an optimal LLM size for this unified scenario and to evaluate the effect of using different LLMs as LGGPT's core implementation, we conduct comparisons across different LLMs with varying scales, including TinyLLaMAv1.1-1.1B \cite{tinyllama2024zhang}, Qwen2-1.5B \cite{qwen22024yang}, Qwen1.5-1.8B \cite{qwen1.52024bai}, NSFW-3B \cite{nsfw}, and LLaMA3-8B \cite{llama32024dubey}. The LLaMA3-8B is fine-tuned with LoRA \cite{lora2022hu} according to the setting described in the LoRA paper. The results are summarized in Table~\ref{Table::comp_llm}. Compared to the equal-sized Qwen2-1.5B and the smaller TinyLLaMAv1.1-1.1B, GPT2-XL performs better on the FID and Alignment metrics, while maintaining competitive performance on Overlap and Max IOU metrics. This overall performance advantage can be attributed to two key factors: (1) \textbf{Data discrepancy.} The layout instruction data differs significantly from the pretraining data of general LLMs, which is heterogeneous data (as described in Sec.~\ref{sec::rw_llm}). Qwen2 and TinyLLaMAv1.1 prioritize broad generality with massive general-purpose instruction data, potentially hampering its adaptability to the specialized layout instructions compared to GPT2. (2) \textbf{Increased architectural complexity.} While recent Qwen2 and TinyLLaMAv1.1 incorporate sophisticated components (GQA \cite{gqa2023ainslie}, Rotary Position Embedding \cite{rope2024su}, RMSNorm \cite{rmsnorm2019zhang}) that benefit general language tasks, GPT2's simpler architecture might be more amenable to fine-tuning on layout-specific tasks. At similar model scales, the increased architectural complexity of recent LLMs could hinder fine-tuning effectiveness on layout generation.

Regarding larger LLMs, \emph{i.e.}, Qwen1.5-1.8B, NSFW-3B, and LLaMA3-8B, a significant decline in performance was observed, particularly with NSFW-3B and LLaMA3-8B. The larger parameter sizes may lead to insufficient optimization across all parameters or potential overfitting, reasonably resulting in degraded outcomes. In addition, the LoRA fine-tuning may not offer sufficient parameter capacity for adapting to the complex, varying layout conditions in unified layout generation. These results suggest that LLMs with $<=$ 1.5B parameters deliver comparable performance without clear discrepancy, while they outperform the larger LLMs by notable margins.

In addition, we compare the inference time cost of these LLMs, in which the testing is consistently conducted using the completion task for fair comparisons. The results are presented in the bottom two lines of Table~\ref{Table::comp_llm}. The results bring forth two surprising findings. (1) The GPT2-XL exhibits the lowest time cost, significantly lower than the equal-size Qwen2-1.5B and even the smaller TinyLLaMAv1.1-1.1B, demonstrating the optimal inference efficiency. (2) Qwen2-1.5B showcases a longer inference time than the larger Qwen1.5-1.8B. These unexpected time costs could be attributed to the architecture discrepancy among LLMs, such as the use of Rotary Positional Embedding, variations in maximum positional encoding lengths, and differences in default vocabulary sizes, which can lead to scenarios where a smaller LLM incurs longer inference times.

Given these observations on model performance and the inference time cost, and considering the potential for future unification of additional layout generation operations, 1.5B could be considered as the optimal parameter size for unified layout generation (implemented with GPT2-XL). Meanwhile, a 1.1B parameter configuration remains a viable secondary option.

\begin{figure*}[t]
	\centering
	\includegraphics[width=\textwidth]{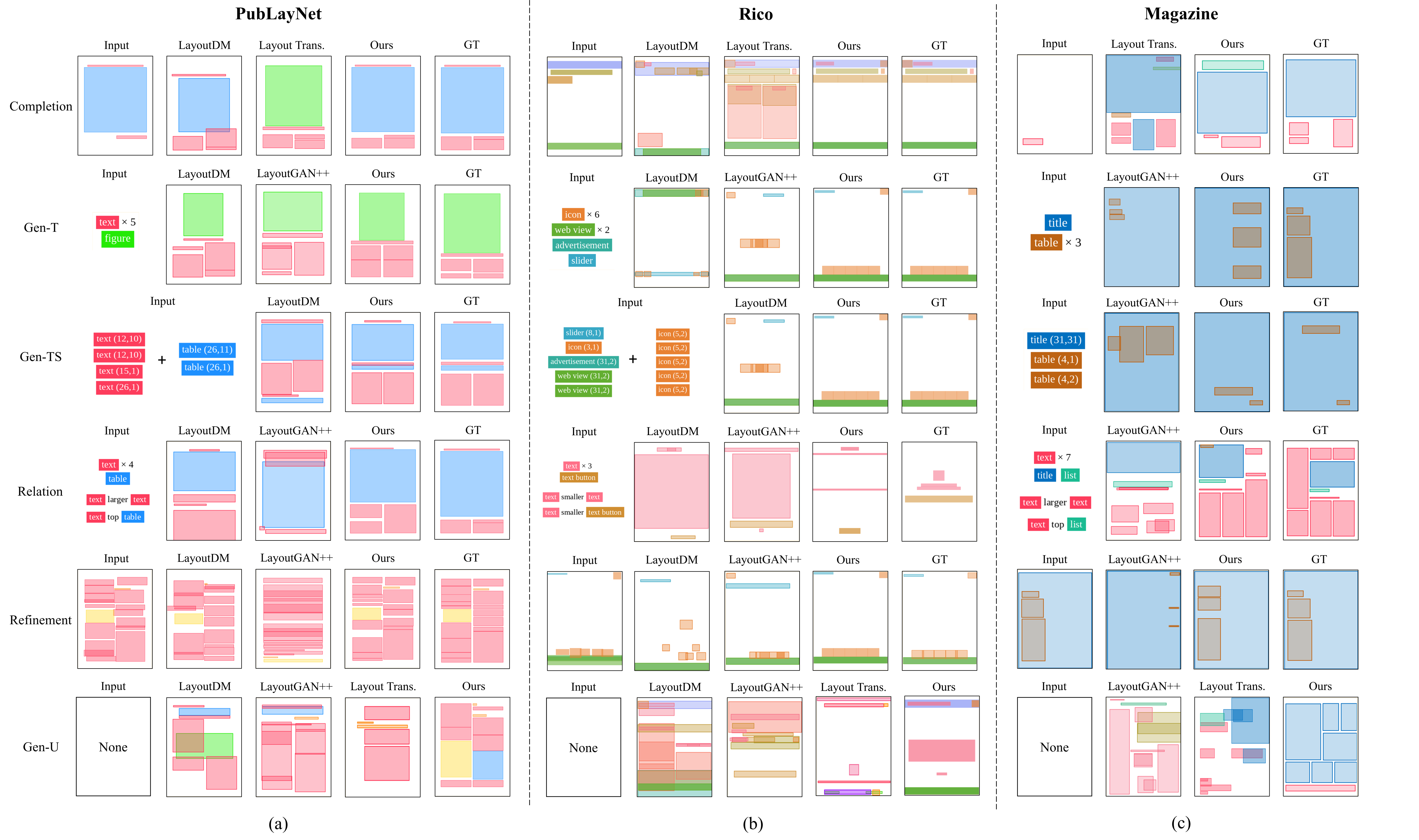}{}
	\caption{Qualitative comparisons with SOTA methods. Zoom in for better view.}
	\label{Fig::visualize}
\end{figure*}

\subsection{Qualitative Results}
Fig.~\ref{Fig::visualize} presents the qualitative comparisons of LGGPT with SOTA methods \cite{layoutdm2023Inoue,layouttf2021gupta,layoutganpp2021kikuchi} across six distinct tasks. It is important to note that these methods are trained with domain-specific data separately, with some being even task-specific. They confront much fewer training difficulties compared to LGGPT, which is designed for unified layout generation. The comparative results depicted in Fig.~\ref{Fig::visualize} (a), (b), and (c) for the PubLayNet \cite{publaynet2019zhong}, Rico \cite{rico2017deka}, and Magazine \cite{magazine2019zheng} datasets, respectively, demonstrate that LGGPT exhibits superior visual fidelity and precise element alignment. Even in a domain-generic scenario, LGGPT manages to fundamentally capture the distribution for each domain of layouts.

We further exhibit visualization examples of LGGPT across different domains of layout data and tasks, as detailed in Fig.~\ref{Fig::pub_visualize} to Fig.~\ref{Fig::slide_visualize} in Appendix~\ref{sec::visualized_prompt}. The results encompass four domains of layout, namely \emph{article}, \emph{App UI}, \emph{magazine}, and \emph{slide}. We can observe that LGGPT generates well-structured and visual-pleasing layouts for specific data types given diverse task conditions. Specifically, on the Gen-UP task, LGGPT showcases its proficiency in translating natural language instructions into high-fidelity layouts. Moreover, we extend to include slide layouts due to their wide adoption in practice, aiming to assess the LGGPT's ability in terms of slide generation. The generation results of slide in Fig.~\ref{Fig::slide_visualize} are also excellent. These proficiencies hint at LGGPT's potential as a versatile tool, serving roles such as a slide designer for streamlined office work or design tasks. However, current training setup with six natural language prompts per layout domain might limit the LLM's breadth of understanding. We can simply incorporate more natural language prompts in training to amplify this functionality.

\section{Potential Functionality Extension}
Since LGGPT is proposed to unify different tasks and domains of layout generation, it is natural to ponder whether additional functionalities could be integrated to broaden its scope of unification. Two potential aspects can be explored.

(1) \textbf{Enhanced Text-to-Layout Generation.} Although LGGPT currently supports text-to-layout operations in the Gen-UP task (Sec.\ref{exp::evaluate}), it primarily uses brief, single-sentence prompts for unconditional generation requirements (see Appendix\ref{sec::nlp_example}). Expanding this to incorporate paragraph-level prompts, similar to the rich text descriptions in \cite{lin2023iccv,layoutprompter2023lin}, could enable more sophisticated conditional layout generation. The enriched prompts could include explicit specifications for element types, positional constraints, and element relationships. Implementing this feature would increase the practicality of LGGPT in text-to-layout generation. (2) \textbf{Content-Aware Layout Generation.} Content-aware layout generation \cite{layoutprompter2023lin,contentaware2024cvpr} refers to considering the content's importance within the layout, ensuring that salient content is not obstructed or overlaid by other layout elements during generation. While LGGPT currently focuses on the unification of content-unaware generation tasks, integrating content-aware generation could further expand the scope of its task unification. Essentially, this functionality can be viewed as the combination of key content identification within the image and the Completion task of layout generation. Given LGGPT's strong performance in the Completion task, it shows promise for adapting to this similar task.

These extensions potentially enhance LGGPT's versatility and applicability across broader layout generation scenarios, especially the content-aware layout generation, which we plan to explore in our future work.

\section{Conclusion}
In this paper, we propose a generic, LLM-based model LGGPT for unified layout generation. We begin with proposing the Arbitrary Layout Instruction (ALI) and Universal Layout Response (ULR) as the standard I/O template. ALI enables LGGPT to execute layout generation for any given domain with arbitrary conditions as input, thus unifying both diverse tasks and distinct domains for the first time in layout generation. We then propose an Interval Quantization Encoding (IQE) strategy to precisely preserve valid layout clues while discarding meaningless placeholders. IQE compresses ALI to a highly condensed structure, fundamentally benefiting the model's comprehension of versatile layouts. To strike a promising proficiency-efficiency balance, we exploit a smaller LLM with 1.5B parameters inside LGGPT. Through instruction tuning based on ALI and ULR, LGGPT is guided to unveil its reasoning prowess for robust unified layout generation.

Experimental results show that LGGPT achieves superior performance and versatility in the demanding domain- and task-generic layout generation, despite having significantly fewer parameters (1.5B) than previous layout generation LLMs (7B or 175B). We further demonstrate the necessity for leveraging LLMs to address this challenging problem. Through comparisons between LLMs of various scales, we reveal that 1.5B could be an appropriate model size for the current scenario, striking an outstanding trade-off between model proficiency and computational efficiency. It is noteworthy that LGGPT is not confined to these four domains, whose potential could seamlessly extend to broader varieties of layout data, such as natural scene layouts \cite{coco2014eccv} and 3D indoor layouts \cite{3dfront2021cvpr}. We hope this work and our findings will facilitate the exploration of LLM for more universal layout generation.

\section*{Data Availability Statement}
The datasets used during and/or analysed in the current study are available in the PubLayNet repository \href{https://developer.ibm.com/exchanges/data/all/publaynet/}{[Link]}, the Rico repository \href{http://www.interactionmining.org/rico.html}{[Link]}, the Magazine repository \href{https://xtqiao.com/projects/content_aware_layout/}{[Link]}, the SPaSe repository \href{https://cvhci.anthropomatik.kit.edu/~mhaurile/spase/}{[Link]}, and the WiSe repository \href{https://cvhci.anthropomatik.kit.edu/~mhaurile/wise/}{[Link]}.

\section*{Declaration}
The authors have no relevant financial or non-financial interests to disclose.

\section*{Acknowledgement}
This research is supported in part by the National Natural Science Foundation of China (Grant No.: 62441604, 62476093).

\bibliographystyle{splncs04}
\bibliography{reference}

\clearpage
\begin{appendices}

\section{Exemplars of Natural Language Prompt for Gen-UP Task}
\label{sec::nlp_example}
In the Gen-UP task, we employ natural language prompts for unconditional layout generation. For each domain of layout data, we predefine six prompt exemplars, consisting of three proprietary and three generic prompts, as listed below:
\vspace{-10pt}
\begin{tcolorbox}[colframe = blue!60!black, colback = blue!3, colupper = black, width=\linewidth, fonttitle = \bfseries, center title, center, title = {Proprietary Prompt Exemplars}]
	\textbf{article}:
	
	\vspace{0.5em}
	1. I need an article layout with various presentation options.
	
	\vspace{0.2em}
	2. Create a clean and organized article layout for a scientific journal article.
	
	\vspace{0.2em}	
	3. Design a professional article layout for a journal.
	
	\tcbline
	
	\textbf{App UI}:
	
	\vspace{0.5em}
	1. Design a highly flexible UI interface for a multi-functional application.
	
	\vspace{0.2em}
	2. Design an intuitive UI interface for a broad user base.
	
	\vspace{0.2em}
	3. Show me a dynamic and diverse UI interface design.
	
	\tcbline
	
	\textbf{magazine}:
	
	\vspace{0.5em}
	1. Please create a versatile magazine layout.
	
	\vspace{0.2em}
	2. I need an informative magazine cover.
	
	\vspace{0.2em}
	3. Design a flexible layout for a magazine publisher.
	
	\tcbline
	
	\textbf{slide}:
	
	\vspace{0.5em}
	1. I want a slide with diverse presentation options.
	
	\vspace{0.2em}
	2. Design an eye-catching slide for a conference presentation.
	
	\vspace{0.2em}
	3. Please generate a slide for content targeted at a wide audience.
\end{tcolorbox}

\begin{tcolorbox}[colframe = blue!60!black, colback = blue!3, colupper = black, width=\linewidth, fonttitle = \bfseries, center title, center, title = {Generic Prompt Exemplars}]
	
	1. Generate a layout of [Layout type].
	
	\vspace{0.2em}
	2. Generate a layout of [Layout type], with [Object number] elements.
	
	\vspace{0.2em}
	3. Generate a layout of [Layout type], with [Object number] elements and [Column number] columns.
\end{tcolorbox}
\noindent \textbf{[Layout Type]} specifies the intended layout type, which should be one of \emph{article}, \emph{App UI}, \emph{magazine}, and \emph{slide}. \textbf{[Object number]} denotes the number of elements. \textbf{[Column number]} represents the number of layout columns. These terms align with the definitions in our proposed prompt-answer template.

\section{Visualization Examples}
\label{sec::visualized_prompt}

\begin{figure*}[ht]
	\centering
	\begin{minipage}[t]{0.45\textwidth}
		\includegraphics[height=0.7\textheight]{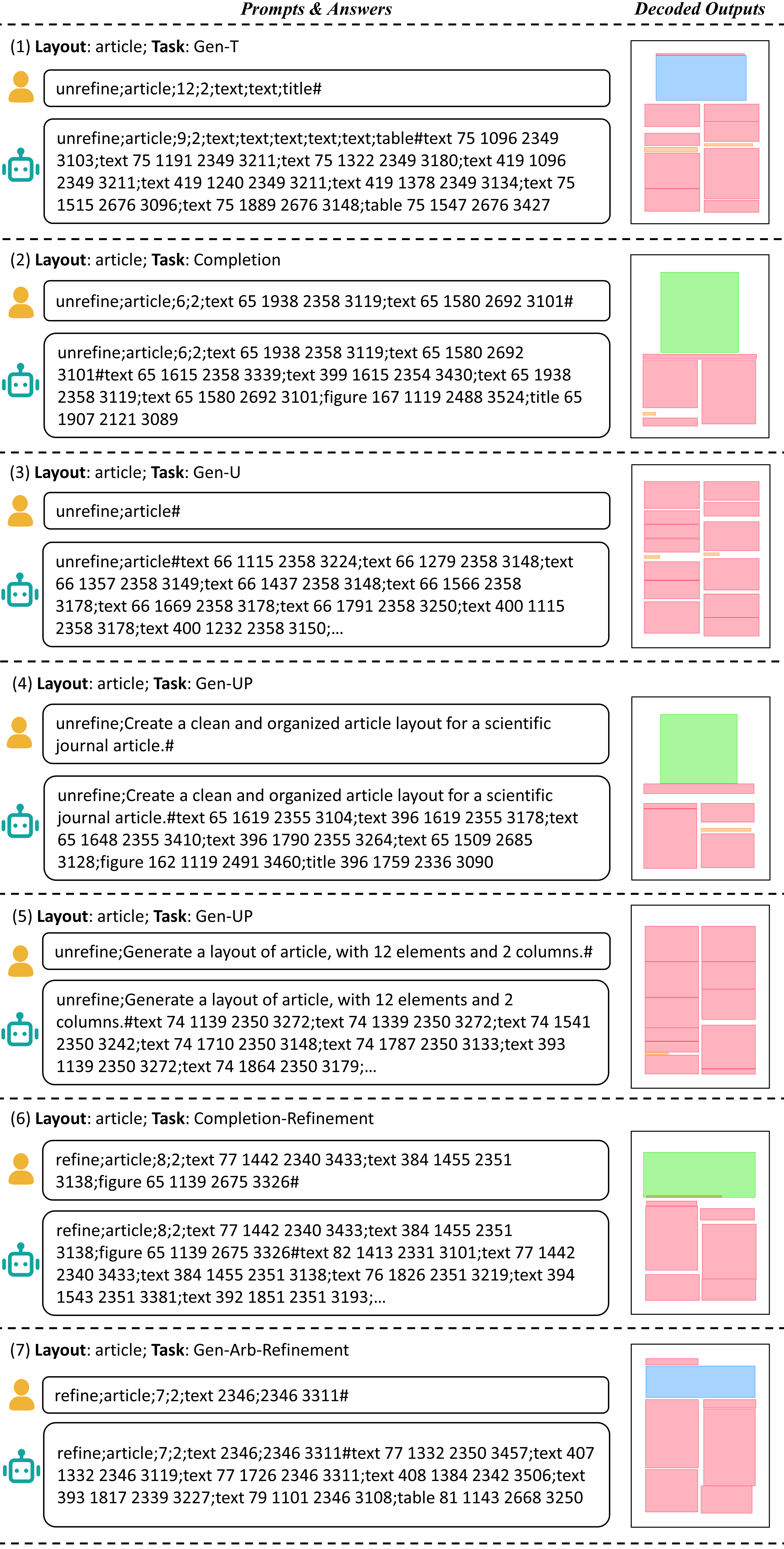}
		\caption{Examples of the generation results of the \emph{article} layout from the proposed LGGPT. Inputs are derived from the PubLayNet \cite{publaynet2019zhong} dataset.}
		\label{Fig::pub_visualize}
	\end{minipage}
	\hfill 
	\begin{minipage}[t]{0.45\textwidth}
		\includegraphics[height=0.7\textheight]{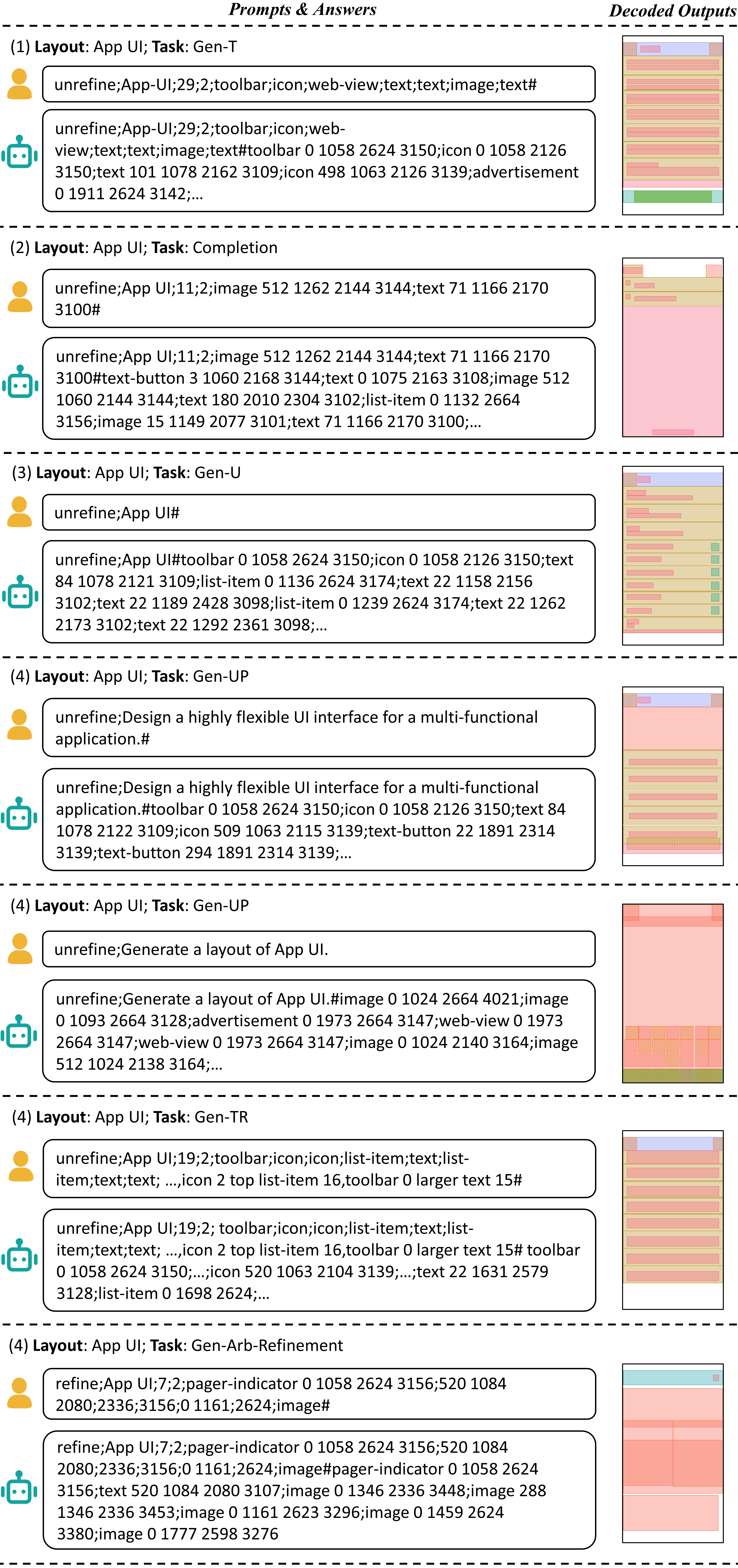}
		\caption{Examples of the generation results of the \emph{App UI} layout from the proposed LGGPT. Inputs are derived from the Rico \cite{rico2017deka} dataset.}
		\label{Fig::rico_visualize}
	\end{minipage}
\end{figure*}

\begin{figure*}[ht]
	\centering
	\begin{minipage}[t]{0.4\textwidth}
		\includegraphics[height=0.65\textheight]{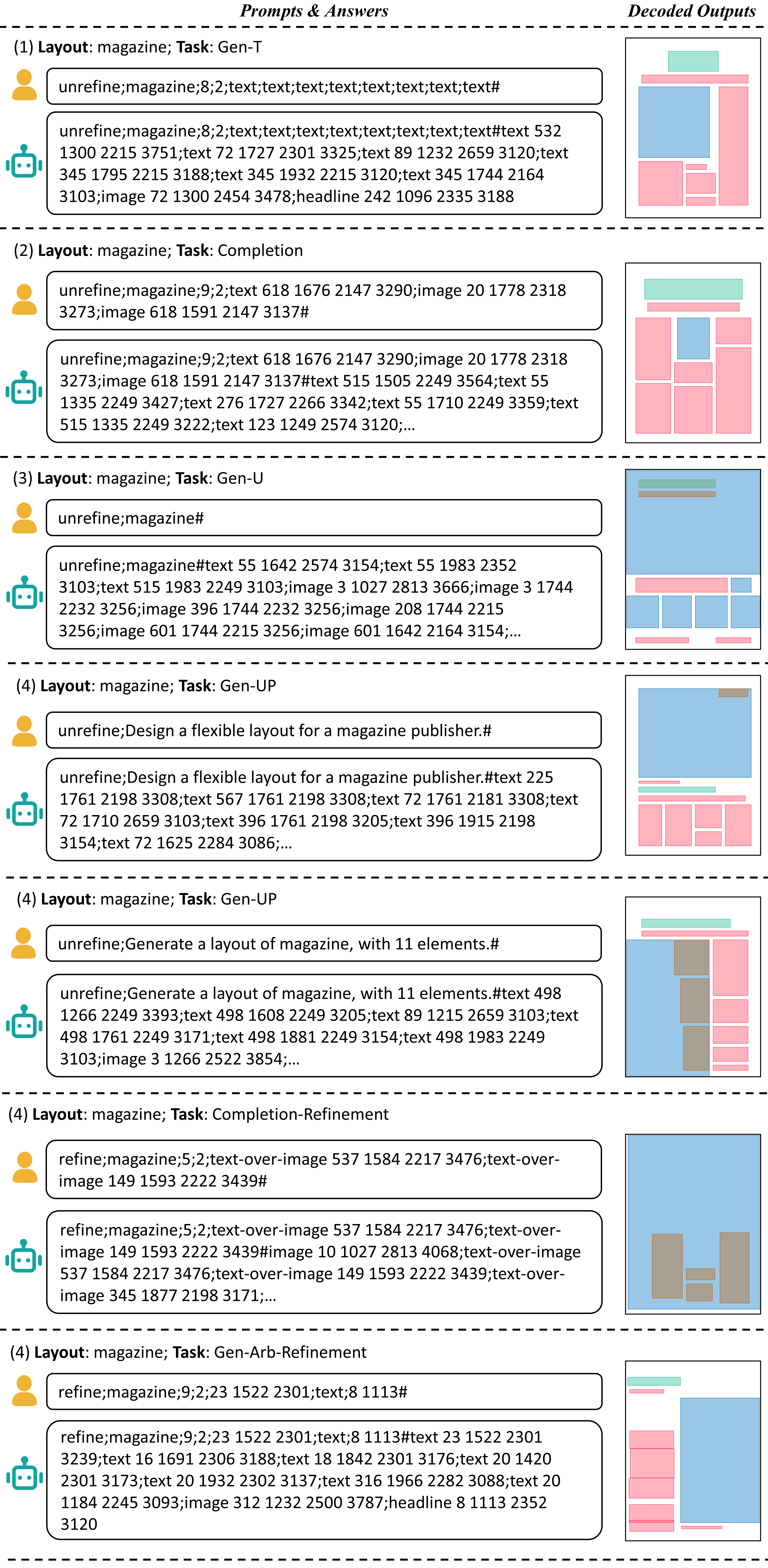}
		\caption{Examples of the generation results of the \emph{magazine} layout from the proposed LGGPT. Inputs are derived from the Magazine \cite{magazine2019zheng} dataset.}
		\label{Fig::mag_visualize}
	\end{minipage}
	\hfill 
	\begin{minipage}[t]{0.5\textwidth}
		\includegraphics[height=0.65\textheight]{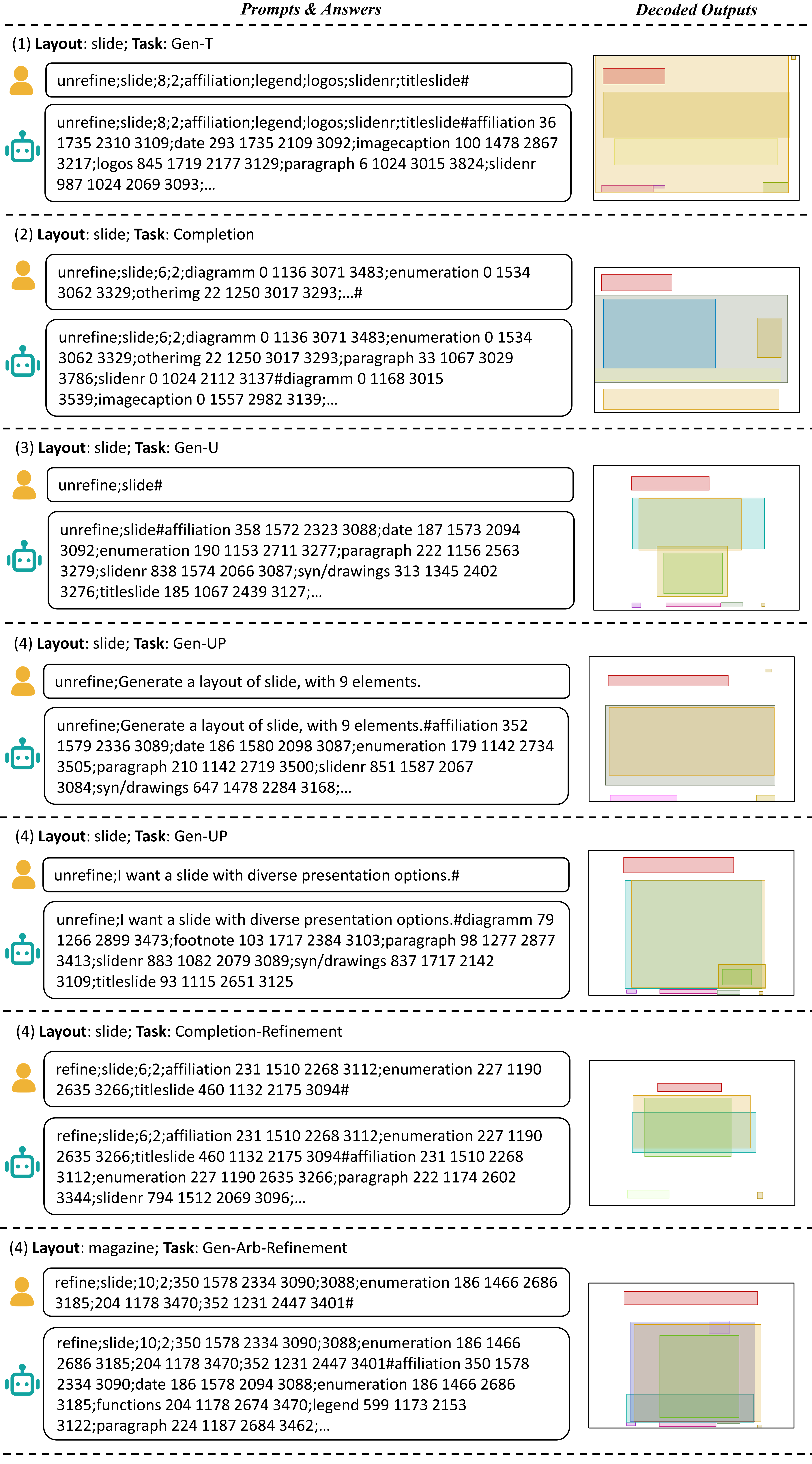}
		\caption{Examples of the generation results of the \emph{slide} layout from the proposed LGGPT. Inputs are derived from the SPaSe \cite{spase2019haurilet} and WiSe \cite{wise2019haurilet} datasets.}
		\label{Fig::slide_visualize}
	\end{minipage}
\end{figure*}

\end{appendices}

\end{document}